\documentclass[acmsmall]{acmart}

\usepackage{makecell}
\usepackage{longtable}
\usepackage{booktabs}
\usepackage{tikz}
\usepackage{mathrsfs}
\usepackage{multirow}
\usepackage{arydshln}
\usepackage{float}
\usepackage{hyperref}
\usepackage{amsmath}
\usepackage{stmaryrd}
\usepackage{latexsym}
\usepackage[edges]{forest}
\usepackage{natbib}

\usetikzlibrary{arrows.meta}
\usetikzlibrary{mindmap}

\AtBeginDocument{%
  }

 \acmJournal{JACM} 

\begin{document}

\title{A Systematic Literature Review on Multi-label Data Stream Classification}

\author{Hiago Freire Oliveira}
\email{h_freire@usp.br}
\orcid{0009-0005-1083-7087}
\affiliation{%
  \institution{Instituto de Ciências Matemáticas e de Computação - ICMC - USP}
  \city{São Carlos}
  \state{São Paulo}
  \country{Brazil}
}

\author{Elaine Ribeiro de Faria Paiva}
\affiliation{%
  \institution{Faculdade de Computação - FACOM - UFU}
  \city{Uberlândia}
  \state{Minas Gerais}
  \country{Brazil}}
\email{elaine@ufu.br}
\orcid{0000-0001-5242-9026}

\author{João Gama}
\affiliation{%
  \institution{University of Porto - INESC TEC}
  \city{Porto}
  \country{Portugal}}
\email{joao.gama@inesctec.pt}
\orcid{0000-0003-3357-1195}

\author{Latifur Khan}
\affiliation{%
  \institution{University of Texas at Dallas - UTD}
  \city{Dallas}
  \state{Texas}
  \country{USA}}
\email{lkhan@utdallas.edu }
\orcid{0000-0002-9300-1576}

\author{Ricardo Cerri}
\affiliation{%
  \institution{Instituto de Ciências Matemáticas e de Computação - ICMC - USP}
  \city{São Carlos}
  \state{São Paulo}
  \country{Brazil}}
\email{cerri@icmc.usp.br}
\orcid{0000-0002-2582-1695}

\renewcommand{\shortauthors}{Freire-Oliveira \textit{et al.}}

\begin{abstract}
Classification in the context of multi-label data streams represents a challenge that has attracted significant attention due to its high real-world applicability. However, this task faces problems inherent to dynamic environments, such as the continuous arrival of data at high speed and volume, changes in the data distribution (concept drift), the emergence of new labels (concept evolution), and the latency in the arrival of ground truth labels. This systematic literature review presents an in-depth analysis of multi-label data stream classification proposals. We characterize the latest methods in the literature, providing a comprehensive overview, building a thorough hierarchy, and discussing how the proposals approach each problem. Furthermore, we discuss the adopted evaluation strategies and analyze the methods' asymptotic complexity and resource consumption. Finally, we identify the main gaps and offer recommendations for future research directions in the field.
\end{abstract}



\maketitle

\section{Introduction}\label{sec1}
Classifying things in the universe is as intrinsic to being a human as reasoning. Aristotle is commonly known to employ some of the first systematic classification of things perceived by the senses. Man has constantly increased the degree of complexity of the classification criteria, so machines have come to hand to help with such tasks, whether by their great speed or by their capability of handling modeling in a mathematically proven fashion.

A machine, in order to classify, must learn the rules that describe such a class. This problem falls under the supervised learning group of algorithms, where the data is used to infer the rules that are previously labeled by a human being (hence being under ``supervision''). One of the first attempts to build a model to learn from data and classify unseen instances was the Perceptron~\cite{Rosenblatt1958}, which improves its performance, that is, its capacity to predict correctly, employing a contour-sensitive projection area. As the number of stimuli the Perceptron learns increases, the probability of recognizing a new stimulus approaches a better-than-chance asymptote.

The Perceptron and many other classification algorithms, such as the classics KNN~\cite{fix1951discriminatory, Cover1967} and Naive Bayes~\cite{Popplestone1971}, treated the learning problem as a one-shot activity, commonly known as \textit{batch} learning. Not much later, Mitchell considered the idea of \textit{rule version space}, a set of rules that are in tune with the training data \cite{Mitchell1977}. His Candidate Elimination algorithm can modify the rule version space in response to new training data, which could be described as \textit{incremental} learning, in opposition to the batch paradigm. A model that was naturally first elected to be modified to an incremental paradigm was the decision tree, given that it can nest concepts indefinitely. One of the first examples is the ID4 algorithm~\cite{Schlimmer1986}, which could get significantly slow given that it is quadratic in relation to the feature space. 

However, as the nature of data shifted more and more to a streaming context nearly the beginning of this millennium, where thousands or millions of records arrive daily at databases in companies, the need for efficient incremental learning algorithms became apparent. An early attempt to address such a problem was proposed by Domingos \& Hulten~\cite{Domingos2000}, establishing one of the landmarks in data stream classification: the Very Fast Decision Tree (VFDT) classifier. The classifier is based on the Hoeffding bound, a statistical measure that aims to decide the number of examples necessary at a node in a tree, given the number of observations, the probability, and the range of a variable.

The research on data stream mining got quite intense with the release of MOA~\cite{Bifet2010}. This Java framework encompasses many tools for implementing and testing online (real-time) machine learning techniques, which soon became the most used pre-built library for data stream published methods. Since then, many problems inherent to the streaming context have been addressed, some deeply, some superficially. These problems include:

\begin{itemize}
    \item High-speed labeling;
    \item High-volume data management;
    \item Dealing with changes in data distribution;
    \item Multi-labeled data;
    \item Dealing with the rise of new labels;
    \item Latency in ground truth label arrival.
\end{itemize}

The first two are perhaps the most investigated issues so far, given that they concern every streaming context. The underlying ideas for solving them are methods with low asymptotic complexity and the ability to handle data in blocks.

Changes in data distribution are a natural issue derived from the impossibility of learning the whole dataset at once. Given that data comes in small chunks, statistical measures are expected to change, which is why the learner must be able to adapt to these changes. This change is formally known as \textbf{concept drift}. In contrast, if the learner must continuously be alert to changes in distribution, it is crucial that it also forgets what was learned in the past, both due to finite storage and outdated concepts.

Multi-label learning rises in response to more realistic learning contexts in batch and streaming, where a data point can have multiple classes simultaneously. However, in a streaming context, the learner may not have access to all possible labels from the beginning, which leads to the need to detect newly emerging labels. This emergence is formally known as \textbf{concept evolution}.

Last but not least, in some conditions, access to the ground truth labels is not feasible at the time. In view of the high-speed data flow and the consequent need for forgetting, as stated above, there can be a latency in the arrival of the labels, which can harm the method's performance if it is too slow compared to the arrival of data points. This issue is the least addressed of all the problems mentioned so far.

In recent years, only one survey on multi-label stream learning came to our attention \cite{Zheng2020}. While it does indeed cover our scope of research, the subject is exceptionally dynamic, with many new methods published since then. Besides, the survey does not cover label latency. Finally, it is not a systematic literature review, as proposed. Surveys or reviews that partially cover our scope were also published. Table~\ref{tab:01} lists the main ones. Our work fills these gaps by conducting a systematic literature review of multi-label classifiers for data streams that addresses the listed~problems.

\begin{table}[ht]
\caption{Recent published surveys and reviews}\label{tab:01}%
\begin{tabular}{|p{4cm}|c|c|c|c|c|c|}
\hline
\makecell{\textbf{Article}} & 
\makecell{\textbf{Year}} & 
\makecell{\textbf{Data} \\ \textbf{stream}} & 
\makecell{\textbf{Multi-label} \\ \textbf{classification}} & 
\makecell{\textbf{Concept} \\ \textbf{drift}} & 
\makecell{\textbf{Concept} \\ \textbf{evolution}} & 
\makecell{\textbf{Label} \\ \textbf{latency}} \\ 
\hline

A survey of multi-label classification based on supervised and semi‑supervised learning~\cite{Han2022} & 2022 & $\bullet$ & X & $\bullet$ & & \\ \hline
Concept Drift Detection in Data Stream Mining: A literature review~\cite{Agrahari2022} & 2021 & X & & X & X & \\ \hline
Data stream classification with novel class detection: a review, comparison and challenges~\cite{Din2021} & 2021 & X & $\bullet$ & X & X & $\bullet$ \\ \hline
A Survey on Multi-Label Data Stream Classification~\cite{Zheng2020} & 2020 & X & X & X & X & \\ \hline
Data stream classification: a review~\cite{Wankhade2020} & 2020 & X & & X & & \\ \hline
A review of improved extreme learning machine methods for data stream classification~\cite{Li2019} & 2019 & X & & X & & \\ \hline
Data stream mining: methods and challenges for handling concept drift~\cite{Wares2019} & 2019 & X & & X & & \\ \hline
An Overview on Concept Drift Learning~\cite{Iwashita2019} & 2018 & X & & X & & \\ \hline
Multi-label Classification: a survey~\cite{STidake2018} & 2018 & & X & & & \\ \hline
A Survey on Ensemble Learning for Data Stream Classification~\cite{Gomes2017a} & 2017 & X & & X & X & \\ \hline
Ensemble learning for data stream analysis: A survey~\cite{Krawczyk2017} & 2017 & X & $\bullet$ & X & & $\bullet$ \\ \hline
\multicolumn{7}{|p{12cm}|}{\textbf{Note:} X = full coverage; $\bullet$ = partial or indirect coverage; empty = not covered.
} \\ \hline
\end{tabular}
\end{table}

This article proposes a systematic literature review of the latest data stream multi-label classification methods. Our contributions can be summarized as follows: i) an exhaustive listing of the asymptotic complexities of most methods; ii) a full hierarchy of all the methods surveyed given their multi-label approach; iii) an exploration of the latest strategies in dealing with label latency, concept drift detection and concept evolution detection; iv) a detailed discussion over the evaluation strategies adopted; v) an identification of the main gaps and future directions for multi-label data stream classification; and vi) recommendations for future directions of research in the field.

The rest of this paper is organized as follows. Section~\ref{sec2} formalizes the underlying concepts to be addressed in reviewing the studies. Section~\ref{meth} presents our research methodology and the retrieved studies. Section~\ref{sec:exploratory} consists of an exploratory analysis of the studies, addressing themes not covered in our research questions. Sections~\ref{sec:q1} to \ref{sec:q6} comprise the core of this review, concerned with answering the proposed research questions. Section~\ref{sec:appl} discusses methods that propose data stream classifiers for specific domains. Finally, in Section~\ref{sec:conclusion}, we present the concluding remarks, the limitations of this work, and proposals for future directions.

\section{Concepts}\label{sec2}

The goal of the multi-label data stream classification task is to learn a model \textit{z} that, given defined metrics, satisfactorily predicts a subset of labels in the set of all possible labels $\mathcal{L}_t$ at time \textit{t} to an instance $x_i$ collected from a stream $\mathcal{DS}$, subject to changes in the underlying distribution $D_t$ at time \textit{t}, subject to the emergence of new classes \textit{g} candidate to being learned and added to $\mathcal{L}_{t+1}$ at time \textit{t+1}. The instance $x_i$ may be accompanied by the ground truth set of associated labels $Y_i \in \mathcal{P}(\mathcal{L})$ or not, so that the model \textit{z} can be updated at time \textit{t+1} with or without labeled instances. In this section, we define each of these concepts individually.

\subsection{Multi-label learning}\label{subsec2.1}

Multi-target learning is a prediction approach in which there is a \textit{set} of output variables that an instance can be associated with \cite{Basgalupp2021, Kocev2013, Xu2019}. So, an instance can be associated up to the number of labels $|\mathcal{L}|$ in any real degree of association to each target $y_i \in \mathcal{Y} = \mathbb{R}^{|\mathcal{L}|}$. It is natural to associate multi-target learning with regression, where the target attribute is numeric.
 
One can introduce constraints to this approach. Multi-label learning is a special case of multi-target learning consisting of the binarization of the target attribute. So, it is natural to associate multi-label learning with classification. Formally, let $\mathcal{D}$ be a multi-label dataset comprised of a \textbf{finite subset} of the cartesian product of $f$  arbitrary sets and the power set of the set of all possible labels $\mathcal{D} = (\mathcal{S}_1 \times \mathcal{S}_2 \times \mathcal{S}_3 \times \dots \mathcal{S}_f \times \mathcal{P}(\mathcal{L}))$, being $f$ the number of features. An instance of such dataset is $\{(x_i, Y_i) \in \mathcal{D} | x_i \in \mathcal{S}_1 \times \mathcal{S}_2 \times \mathcal{S}_3 \times \dots \mathcal{S}_f, Y_i \in \mathcal{P}(\mathcal{L})\}$, so that $x_i$ is an element of the input space $\mathcal{X}$, and $Y_i$ a vector of the output space $\mathcal{Y}$ . The length of $x_i$ is the number of features $f$, and the length of $Y_i$ is $|\mathcal{L}|$. Hence, a multi-label classifier is a function that maps a vector of features to a vector of predicted labels $\mathcal{C}:\mathcal{X} \rightarrow \mathcal{Y}$ \cite{Herrera2016-multi}.

A second constraint, not convenient in this study, is, besides binarizing the target attribute (therefore, a classification problem), to limit the return of the classifier to a single class out of the \textit{set} of available classes \cite{grandini2020metrics}. This constraint is called multi-class learning, and it can be perceived as a special case of the multi-label classification, where $y_i \in \mathcal{L}$, being $\mathcal{L}$ the set of all possible labels, as mentioned.

\begin{figure}
\begin{forest}
  for tree={
    align=center,
    font=\sffamily\scriptsize,
    edge+={thick},
    fork sep=3pt,
    s sep=0.08mm,
    l sep=3mm,
    grow=east,
    scale=1,
  },
  forked edges,
  [Approaches,
    [Ensemble,
        [Heterogeneous
            [EML]
            [MULE]
        ]
        [Homogeneous
            [Hierarchic
                [RFML-C4.5]
                [RF-PCT]
            ]
            [Multi-class
                [RAkEL]
                [EPS]
            ]
            [Binary
                [AdaBoost.MH]
                [ECC]
                [EBR]
            ]
        ]
    ]
    [Algorithm\\Adaptation,name=AA
        [kNN
            [kNNc]
            [ML-kNN]
        ]
        [Support vector machines
            [Rank-SVM]
        ]
        [Neural networks
            [ML-RBF]
            [BP-MLL]
        ]
        [Trees
            [ML-Tree]
            [PCT]
            [ML-C4.5]
        ]
    ]
    [Problem\\Transformation,name=PT
        [Multiple labels
            [HOMER]
            [PS]
            [CC]
            [LP]
        ]
        [Pair of labels
            [QWML]
            [CLR]
            [RPC]
        ]
        [Single Label
            [BR+]   
            [2BR]
            [BR]
        ]
    ] 
    ]
\end{forest}
\caption{Hierarchy of multi-label classification methods. Adapted from \cite{Herrera2016-multi, Herrera2016-adaptation, Herrera2016-transformation, STidake2018, Guehria2023, Madjarov2012}.}
\label{fig:1}
\end{figure}
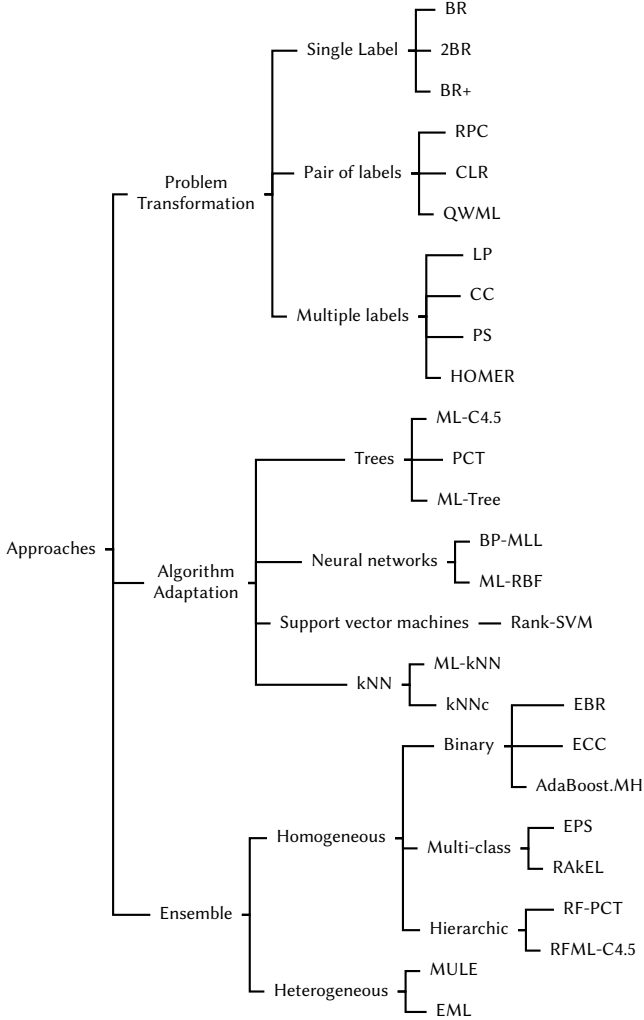

Three approaches are classically listed to develop a multi-label classifier: problem transformation, algorithm adaptation \cite{ganda2018survey}, and ensemble \cite{Guehria2023}. A corresponding  hierarchy is proposed in Figure~\ref{fig:1}.


\subsubsection{Problem transformation}\label{pt}

This approach involves mapping the posed multi-label problem  to a set of atomic-label tasks in an algorithm-independent fashion \cite{ganda2018survey, Cherman2011, Tsoumakas2007-overview, Herrera2016-transformation}. The most widely used transformation is the Binary Relevance (BR), in which a multi-label dataset, as defined in \ref{subsec2.1}, is transformed in a set of $|\mathcal{L}|$ single-label datasets $\{\mathcal{D}_1 = (\mathcal{S}_1 \times \dots \times \mathcal{S}_f \times \mathcal{L}_1), \mathcal{D}_2 = (\mathcal{S}_1 \times \dots \times \mathcal{S}_f \times \mathcal{L}_2), \dots, \mathcal{D}_{|\mathcal{L}|} = (\mathcal{S}_1 \times \dots \times \mathcal{S}_f \times \mathcal{L}_{|\mathcal{L}|})\}$, addressed independently by single-label classification algorithms. The output prediction is the union of all the independent positive results. The main advantage is the low complexity, but its drawback is the loss of the relationship between labels. To handle this issue comes Classifier Chains (CC), a flavor of binary relevance in which BR classifiers are linked in a chain where a link's feature space is broadened with the binary associations of all the links \cite{Read2009}. Besides are the 2BR \cite{Tsoumakas2009CorrelationBasedPO} and BR+ \cite{AlvaresCherman2012}, which also intend to address label dependency while still employing BR.

Another transformation is the Label Power Set (LP), in which the multi-label dataset is transformed into a multi-class dataset, such that the label space is no longer the set of all possible labels but the set of all \textbf{existing combinations} of labels in the original dataset. Hence, let $\mathcal{L}$ be the original label space $\mathcal{L} = \{y_1, y_2,y_3,\dots,y_{|\mathcal{L}|}\}$, the transformed label space $\mathcal{L}' = \{y_{1,2,3}, y_{1,3}, y_{2},\dots\}\}$, so that $|\mathcal{L}'|$ is up to $|\mathcal{P}(\mathcal{L})|$. The main advantage is the preservation of the relationship between labels, but its drawback is the possible exponential increase in the dimension of the dataset, leading to many classes with few positive examples \cite{Cherman2011}.

In Pairwise classification (PC), the dataset in transformed into a set of $\binom{|\mathcal{L}|}{2} = k$ single-label datasets, in which each new label is a pair of original labels $\{\mathcal{D}_1 = (\mathcal{S}_1 \times \dots \times \mathcal{S}_f \times (y_1,y_2)), \mathcal{D}_2 = (\mathcal{S}_1 \times \dots \times \mathcal{S}_f \times (y_1,y_3)), \dots, \mathcal{D}_{|\mathcal{L}|-1} = (\mathcal{S}_1 \times \dots \times \mathcal{S}_f \times (y_1,y_{|\mathcal{L}|})), \dots, \mathcal{D}_k = (\mathcal{S}_1 \times \dots \mathcal{S}_f \times (y_{|\mathcal{L}|-1},y_{|\mathcal{L}|}))$, such that only instances that hold one of the labels, but not both, are kept. Then, the classification proceeds by choosing the label with the highest votes in a ranking \cite{Yapp2020, Hllermeier2008}. In order to allow its use in multi-label scenarios, the Calibrated Label Ranking (CLR) poses an additional label that acts as a splitting point between the relevant and irrelevant labels in the same ranking \cite{Frnkranz2008}. It keeps the relationship between labels and sticks to binary subproblems.

Worthy of notice is the Pruned Sets (PS) method, which prunes away rare label sets and reintroduces them as more frequent subsets \cite{Read2008}; the Quick Weighted Multi-label Learning (QWML) \cite{LozaMenca2010}, an extension of CLR to increase efficiency that stops the ranking calculation when the splitting point can no longer change; and the Hierarchy Of Multi-label classiﬁERs (HOMER) \cite{Tsoumakas2008EffectiveAE}, a divide-and-conquer approach that splits the label space in a tree of classification tasks with fewer labels, followed by propagation of labels from the root to the children.

\subsubsection{Algorithm adaptation}

This approach consists of selecting an existing machine learning method and adapting, extending, and customizing it to a scenario with several outputs \cite{Madjarov2012, Herrera2016-adaptation}. A simple yet powerful method is the Multi-Label k-Nearest Neighbors (ML-kNN) \cite{Zhang2007}, which extends the ubiquitous kNN to multi-label scenarios. It works by outputting a category vector built through a \textit{maximum a posteriori} estimate, such that the splitting happens as a BR classifier \cite{Herrera2016-adaptation} and the neighbors are found similarly as the original kNN algorithm \cite{Cover1967}. 
An extension is the kNNc, which incorporates prototype selection to reduce the training set, limiting the prediction to the labels learned but using the whole data set \cite{CalvoZaragoza2015}. 

Multi-label C4.5 (ML-C4.5) \cite{Clare2001}, Predictive Clustering Trees (PCT) \cite{blockeel2000topdown} and ML-Tree \cite{QingyaoWu2015} adapt the tree-based logic to multi-label. ML-C4.5 adapts the authoritative C4.5 \cite{Quinlan1993} by modifying the entropy computation to sum the individual labels' entropies so that the leaves can bear multiple labels. PCT works by inducing a clustering tree in a top-down fashion, such that a prototype function yields a vector of labeling probabilities for a given example, and Gini indices give the variance. Finally, ML-Tree employs one-versus-all SVM classifiers at the tree levels and propagates labels through nodes hierarchically.

Neural network methods include the Backpropagation for Multi-label Learning (BP-MLL) \cite{MinLingZhang2006} and the Multi-Label Radial Basis Function (ML-RBF) \cite{Zhang2009}. BP-MLL is a label ranking-based algorithm that introduces an error function designed for multi-label scenarios, which is minimized by gradient descent mixed with backpropagation. In turn, ML-RBF defines the prototype vector of $|\mathcal{L}|$ RBFs through $|\mathcal{L}|$ clustering executions in the samples associated with each label, followed by a weighting learned by minimization of an error function.

Ranking Support Vector Machine (Rank-SVM) \cite{Elisseeff2002} employs the minimization of Ranking Loss with maximization of the margin to train an SVM-like model, followed by adjustment of the cutting threshold to draw out the relevant labels.

\subsubsection{Ensemble}

This approach combines multiple multi-label classifiers' outputs to improve generalization and reduce overfitting in yielding a final sub-label set \cite{Guehria2023, Moyano2018}. The composition of these classifiers can be split into (i) \textit{homogeneous} and (ii) \textit{heterogeneous}, being the first case when the base classifiers combined are of the same type and the latter of different types. Plus, the combination strategy can be split according to the number of models used: (a) \textit{fusion} (all), (b) \textit{selection} (one), and (c) \textit{ensemble pruning} (subset). Furthermore, an ensemble is \textit{global} if all labels use the same model combination, or \textit{local} if each label possesses its own combination \cite{Papanikolaou2017}.

Immediate consequences of the aforementioned problem transformation techniques are Ensemble of Binary Relevance (EBR) \cite{Read2011}, Ensemble of Classifier Chain (ECC) \cite{Read2009} and Ensemble of Pruned Sets (EPS) \cite{Read2008}. As an extension of the Label Power Set approach, Random k-Labelsets (RAkEL) takes into account a small random label subset and then models a single-label classifier to predict each element in the power set of such subset~\cite{Tsoumakas2007}.

Boosting methods combine several weak classifiers sequentially, aiming to reduce bias. AdaBoost.MH \cite{Schapire1999} is a multi-label boosting technique based on decomposing the problem into binary classification problems and averaging their error rate to minimize Hamming Loss.

Random Forest of Predictive Clustering Trees (RF-PCT) \cite{Kocev2007} and Random Forest of Multi-Label C4.5 (RFML-C4.5) \cite{Clare2001} are immediate consequences of some of the preceding tree-based algorithm adaptation techniques, as the random forest is a genre of ensemble of tree-structured classifiers that vote unitarily for a class \cite{Breiman2001}.

Finally, heterogeneous methods to be listed are MUlti-Label Ensemble (MULE) \cite{Papanikolaou2017}, which elects, via McNemar test, the suitable model among the members for each distinct label, and Ensemble of Multi-Label classifiers (EML) \cite{Tahir2012}, that proposes five strategies of combination of different classifiers to deal both with sample imbalance and label correlation issues.

\subsection{Data stream}\label{subsec2.2}
A data stream can be defined as a continuous flow of data \cite{AlKhateeb2012}, a read-only sequence of data \cite{10.7551/mitpress/10654.003.0009}, or even a stochastic process of independent and continuous events \cite{Gama2010_2}. Data arrives in chunks. Let $C_n = x_{(n-1)L+1},\dots,x_{nL}$ be a chunk of data, where $x_i$ is the $i$th instance in the stream, L is the chunk length, and $C_n$ is the latest data chunk \cite{AlKhateeb2012}. Then, a data stream $\mathcal{DS} = \{C_1, C_2, C_3,\dots,Cj\}$ is a set of chunks $C_t$ (or examples, if L=1) available at time $t \in \mathcal{T}_x = \{1,2,3,\dots,j\}$. The data stream is potentially unbounded and infinite as $\lim_{j\to\infty}$.

In a multi-label data stream classification context, each $x_i$ data point has a label set $Y_i \in \mathcal{P}(\mathcal{L})$ available at time $\textit{t} \in \mathcal{T}_Y$, where $\mathcal{T}_Y$ depends on the context, as exposed in \ref{subsec2.3}. Let $\mathcal{L_t} = \{l_{1_{t}},\dots,l_{C_{t}}\}$ be the set of all possible classes an instance $x_i$ can be associated with, that is, the learned and retained classes. A class \textit{g} is novel if $g \notin \mathcal{L}$, but an instance $x_i$ might be labeled as pertaining to \textit{g}. The novel class remains novel until the model is trained again and adds \textit{g} to the set of possible labels $\mathcal{L} = \mathcal{L} \cup \{g\}$.

\subsection{Label latency}\label{subsec2.3}

Label latency is the difference between an example's arrival timestamp $t_{x} \in \mathcal{T}_x$ and its respective label set's arrival timestamp $t_{Y} \in \mathcal{T}_Y$ in a classification context. Let the delay $d = t_{Y} - t_{x}$. Thus, one can derive three types of label latency according to the nature of $d$ \cite{Gomes2017}: 
\begin{itemize}
    \item \textbf{Immediate.} If $d=0$, the timestamps of the labels are synchronized with the timestamps of the instances,  i.e. $\mathcal{T}_y = \mathcal{T}_x$, so that the learner has access to them before the next instance arrives;
    \item \textbf{Delayed.} If $d>0$, labels arrive with delay, i.e., $\mathcal{T}_y \ne \mathcal{T}_x$, so that the learner does not have access to them before the next instance arrives, but might have access in feasible time. Notably, the delay mechanism can be deterministic $d = c$, being $c$ a constant, or stochastic $d \backsim D_t$, being $D_t$ a probability distribution \cite{Plasse2016};
    \item \textbf{Infinite.} If $d>0$ and $\lim_{d \to \infty}$, labels are never available to the learner, i.e $\mathcal{T}_y \ne \mathcal{T}_x$ and $\mathcal{T}_y = \emptyset$. It is a special case of delay.
\end{itemize}

\subsection{Non-stationary data streams}\label{novelty}

Given the non-stationary nature of data streams, they are subject to two main phenomena: \textit{concept drift} and \textit{concept evolution}, elucidated as follows.

\subsubsection{Concept drift}\label{drift}

A concept drift is a change in the underlying distribution of a data stream over time \cite{Agrahari2022}. Formally, let $C_t$ be the $t$th chunk (or instance) in a data stream, generated from a source with an underlying distribution $D_t$ available at time $t \in \mathcal{T}_x = \{1,2,3,\dots,j\}$. A concept drift happens when $D_{t} \neq D_{t+1}$ given the arrival of the chunks $C_{t}$ and $C_{t+1}$ \cite{Faria2015, Farid2013}.

The concept drift detection strategies can be understood as passive or active. Passive strategies consist of not employing any explicit drift detector, but with a continuous update to handle the changes. Active strategies do employ an explicit drift detector (e.g., Adwin), from where the model may begin the procedures to learn these changes \cite{Heusinger2020}.

It is possible to evaluate the drift in terms of which probability distribution changes \cite{Agrahari2022}: (a) real drift, which refers to change in posterior probability $p(l_i \in \mathcal{L}|X)$; (b) virtual drift, which refers to change in the class conditional probability $p(X|l_i \in \mathcal{L})$. Moreover, one can evaluate the drift in terms of the degree of change \cite{Gama2014}:
\begin{itemize}
    \item Incremental: comprises intermediate concepts;
    \item Sudden/Abrupt: does not comprise intermediate concept;
    \item Gradual: does not comprise intermediate concepts, but both concepts coexist briefly;
    \item Recurring: return of a past concept.
\end{itemize}

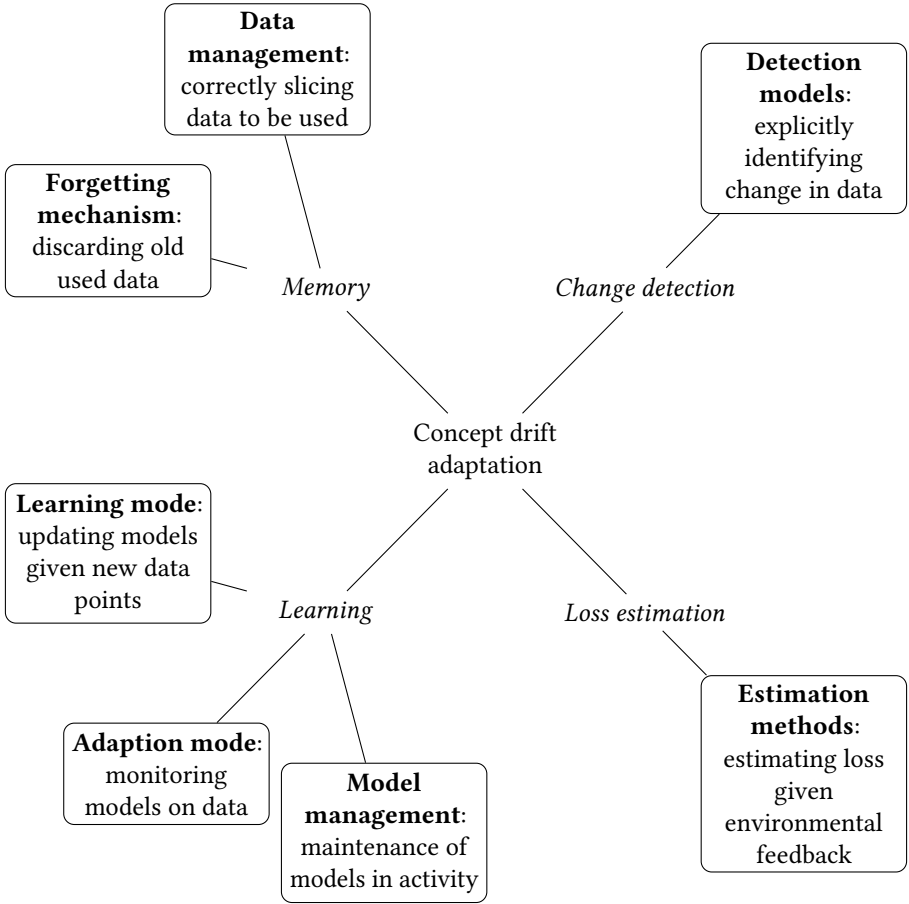
\begin{figure}[ht]
\begin{tikzpicture}[grow cyclic, text width=2.5cm, align=flush center,
	level 1/.style={level distance=3cm,sibling angle=90},
	level 2/.style={level distance=3cm,sibling angle=60}]
    
\node{Concept drift adaptation}
child { node {\textit{Learning}}
	child { node[draw, rectangle, rounded corners] {\textbf{Learning mode}: updating models given new data points}}
 	child { node[draw, rectangle, rounded corners] {\textbf{Adaption mode}: monitoring models on data}}
  	child { node[draw, rectangle, rounded corners] {\textbf{Model management}: maintenance of models in activity}}
}
child { node {\textit{Loss estimation}}
	child { node[draw, rectangle, rounded corners] {\textbf{Estimation methods}: estimating loss given environmental feedback}}
}
child { node {\textit{Change detection}}
	child { node[draw, rectangle, rounded corners] {\textbf{Detection models}: explicitly identifying change in data}}
}
child { node {\textit{Memory}}
	child { node[draw, rectangle, rounded corners] {\textbf{Data management}: correctly slicing data to be used}}
        child { node[draw, rectangle, rounded corners] {\textbf{Forgetting mechanism}: discarding old used data}}
};
\end{tikzpicture}
\caption{Concept drift adaptation comprehended in four non-exclusive modules and what their respective dimensions are concerned with. Adapted from \cite{Gama2014}.}
\label{fig:2}
\end{figure}

There are many approaches to building a concept drift method to detect these changes. These methods can be dissected into four modules, as illustrated in Figure~\ref{fig:2}. The memory module splits into two submodules: data management (given the impossibility of dealing with the whole stream at once) and forgetting mechanism (given the impossibility of storing the arriving data points indefinitely). The change detection module refers to the techniques for recognizing change in the stream (active). The loss estimation consists of the feedback system of ground truth and comparison with the prediction. The learning module splits into three sub-modules: learning mode (update when new data arrives), adaptation mode (monitoring model to changes - passive), and model management (assessing the better models - a bigger concern for ensemble methods).

\subsubsection{Concept evolution}\label{evol}

Concept evolution is the rise of new classes in the stream of data \cite{Masud2010, Faria2015}. A class is a \textit{novel class} if none of the classifiers of the model were trained with it in the \textit{offline} phase \cite{Gaudreault2024}. Formally, let $\mathcal{L}^{off} = \{\ell_{tr_{1}}, \ell_{tr_{2}},\dots,\ell_{tr_{|\mathcal{L}|}}\}$ be the set of labels known in the training phase. Then, $\mathcal{L}^{on} = \mathcal{L}^{off} \cup {\{\ell_{novel_{1}}, \ell_{novel_{2}}, \dots, \ell_{novel_{k}}\}}$ is the new set of labels, expanded with the \textit{k} novel classes detected as new data arrive in the \textit{online} phase.

Reciprocally, if a class does not have any positive instances at a given time, the class can be totally or partially forgotten to save memory. If an instance with this very label appears again, this rising class is considered a recurring class.

\subsection{Evaluation}\label{subsec2.6}

Evaluating a data stream involves many intrinsic issues, as the classifiers must self-modify their internal state throughout the data flow so that there is no unitary generalization to be judged. Beyond \textit{generalization power}, \textit{space} and \textit{learning time} are more prominent matters in streaming, as they can directly affect the prediction of the following data point \cite{Gama2012}.

In a multi-label classification task, given an instance $x_i$, the outcome is a set of predicted labels $\hat{Y}_i\in  \mathcal{P(L \cup \{g\})}$ that might pertain to the known labels $\mathcal{L}$ or be pointed out as a new class $g$. In a supervised context, the instance may come along (instantly or delayed) with the ground truth set of labels $Y_i\in  \mathcal{P(L \cup \{g\})}$. The primary metrics consider comparing these sets in order to assess the performance of the classifier, as one can see next.

\subsubsection{Metrics}\label{ev1}

The performance of a multi-label classification task can be assessed by many distinct metrics, which can be grouped into example-based, label-based, ranking-based, and efficiency measures \cite{Xu2019, grandini2020metrics, STidake2018, Tsoumakas2007-cc}. The most relevant metrics are listed in Table~\ref{tab:02}.

\paragraph{Example-based}\label{example-based}

This group consists of methods that evaluate the performance of multi-output learning models concerning each data instance. After being separately addressed, the overall performance is yielded by the mean of all individual results.

Precision $P$ is the ratio between true positives and the positive predictions along all labels. Recall $R$ is the ratio between true positives and the actual positive examples along the labels. $F_1$ is a metric that mixes precision and recall by computing their harmonic mean. Accuracy $acc$ is the ratio between true positives and the sum of positive predictions and actual positives.

Hamming loss ($HL$) is concerned with example-label pairs that are not correctly predicted, where $\mathcal{I}$ is the interpretation of the inequality test. Subset accuracy ($SA$), on the other hand, concerns the correctly predicted instances.

\paragraph{Label-based}\label{label-based}

This group consists of metrics that evaluate the performance regarding each label covering the examples, which are then averaged. Precision ($P$), recall ($R$), and $F_1$ are defined the same as in example-based, but the averaging strategy is pertinent to the final value.

The averaging can be done in a macro or micro fashion. In macro ($M$), labels are equally weighted, such that each is independently evaluated and, subsequently, the average is obtained over them all. In micro ($m$), observations are equally weighted, disregarding possible differences between labels.

\paragraph{Ranking-based}

This group consists of metrics that evaluate the model, taking into consideration the correct ordering of the predicted labels.

The average precision is the share of labels ranked higher than a given label over all true labels, where $Y_i$ is the true label set.

One-error is the average, over all instances, of how much a top-ranked label does not appear in $Y_i$, where $\mathcal{I}$ is the interpretation of the inequality test and $f(x_i,y)$ is the classification function. 

Finally, ranking loss is the average of the amount of label pairs that are ordered in a reversed fashion, where $\bar{Y}_i$ is the complementary set of irrelevant pairs.

\begin{table}[ht]
\caption{Multi-label evaluation metrics}
\begin{center}
\begin{tabular}{c|c|c|c}
    \toprule
    \multicolumn{2}{c}{}&\(\displaystyle P = \frac{1}{p}\sum_{i=1}^{p} \frac{TP_i}{TP_i + FP_i} \)&\(\displaystyle acc = \frac{1}{p}\sum_{i=1}^{p} \frac{TP_i}{TP_i + FP_i + FN_i} \)\\
    \multicolumn{2}{c}{\multirow{1}{*}{Example-based}}&\(\displaystyle R = \frac{1}{p}\sum_{i=1}^{p} \frac{TP_i}{TP_i + FN_i} \)&\(\displaystyle HL = \frac{1}{pq}\sum_{i=1}^{p} \sum_{j+1}^{q} \mathcal{I} [\hat{y}_{j}^{(i)} \neq y_{j}^{(i)}] \)\\
    \multicolumn{2}{c}{}&\(\displaystyle F_1 = 2 \times \frac{P \times R}{P + R} \)&\(\displaystyle SA = \frac{1}{p}\sum_{i=1}^{p} \mathcal{I} [\hat{y}^{(i)} = y^{(i)}] \)\\
    \hline
    \multicolumn{2}{c}{}&\(\displaystyle M_P = \frac{1}{q}\sum_{j=1}^{q} \frac{TP_j}{TP_j + FP_j} \)&\(\displaystyle m_P = \frac{\sum_{j=1}^{q}TP_j}{\sum_{j=1}^{q}TP_j + \sum_{j=1}^{q}FP_j} \)\\
    \multicolumn{2}{c}{\multirow{1}{*}{Label-based}}&\(\displaystyle M_R = \frac{1}{q}\sum_{j=1}^{q} \frac{TP_j}{TP_j + FN_j} \)&\(\displaystyle m_R = \frac{\sum_{j=1}^{q}TP_j}{\sum_{j=1}^{q}TP_j + \sum_{j=1}^{q}FN_j} \)\\
    \multicolumn{2}{c}{}&\(\displaystyle M_{F_{1}} = \frac{1}{q}\sum_{j=1}^{q} \frac{2TP_j}{2TP_j + FP_j + FN_j} \)&\(\displaystyle m_{F_{1}} = 2 \times \frac{m_p \times m_r}{m_p + m_r} \)\\
    \hline
    \multicolumn{2}{c}{}&\multicolumn{2}{c}{\(\displaystyle average~precision = \frac{1}{p}\sum_{i=1}^{p} \frac{1}{|Y_i|} \sum_{y \in Y_i} \times \frac{|\{y' | rank_{f}(x_i, y') \leq rank_{f}(x_i,y), y' \in Y_i\}|}{rank_{f}(x_i,y)} \)}\\
    \multicolumn{2}{c}{\multirow{1}{*}{Ranking-based}}&\multicolumn{2}{c}{\(\displaystyle one-error = \frac{1}{p}\sum_{i=1}^{p} \mathcal{I}[[arg~max f(x_i,y)] \notin Y_i] \)}\\
    \multicolumn{2}{c}{}&\multicolumn{2}{c}{\(\displaystyle ranking~loss = \frac{1}{p}\sum_{i=1}^{p} \frac{1}{|Y_i||\bar{Y}_i|} |\{(y', y'') | f(x_i, y') \leq f(x_i, y''), (y', y'') \in Y_i \times \bar{Y}_i\}| \)}\\
    \hline
\end{tabular}
\end{center}
\label{tab:02}
\end{table}

\subsubsection{Procedures}\label{ev2}

There are two pervasive procedures to evaluate a data stream in a classification context: holdout and prequential~\cite{Gama2012, Gama2014}.

The holdout procedure is akin to holdout in batch learning, as the current model is regularly applied to a test set. However, the holdout is often unfeasible, as a holdout set in streaming must represent a context in a given time, and one cannot always know which examples are comprised by the concept active at the same time.

On the other hand, \textit{prequential} (predictive sequential) or \textit{interleaved test-then-train} \cite{Bifet2010} deals with individual instances and tests the model with each example prior to using it for training, updating accuracy incrementally as an example is used to test. A prequential evaluation may make use of three strategies: (i) a landmark window, (ii) a sliding window, or (iii) a forgetting mechanism. The error is the accumulated sum of a loss function, at time \textit{t}, between the predicted and ground truth values, hence: $P_e (t) = \frac{1}{t}\Sigma_{k=1}^{t}loss(y_k,\hat{y}_k)$.

\section{Methodology} \label{meth}

This section is concerned with the method employed to conduct this Systematic Literature Review (SLR). This SLR is structured according to the framework by Kitchenham and Charters \cite{Kitchenham2007} and Petticrew and Roberts \cite{Petticrew2006}. First, we elaborated on research questions to define the scope of the study. Second, we established the search criteria by choosing the data sources and building the search string by means of the PICOC criteria, which stands for Population, Intervention, Comparison, Outcome, and Context. Then, we listed the criteria for inclusion and exclusion, followed by the study selection and composition of the data extraction form. More details are provided in the following sections.

\subsection{Research Questions}\label{rese}
Given the problems listed in Section~\ref{sec1}, six research questions could be conceived, as exposed in Table~\ref{tab:03}.

\begin{table}[h]
\caption{Research questions}\label{tab:03}%
\begin{tabular}{@{}llll@{}}
\toprule
 & Question\\
\midrule
Q1 & How does the classifier work? \\
Q2 & How does it handle label latency? \\
Q3 & What is the scheme to handle concept drift detection?\\
Q4 & What is the scheme to handle concept evolution detection?\\
Q5 & What is the evaluation strategy? \\
Q6 & What are the limitations and proposals for future work?	 \\
\bottomrule
\end{tabular}
\end{table}

The first question \textbf{Q1} is a broad question that refers mainly to the multi-label classification task (Sections \ref{subsec2.1}, \ref{subsec2.2}) itself and the pipeline of algorithms according to the number of problems the method aims to deal with. Moreover, as some problems impose greater costs and the methods cannot be compared equally, this question also takes into account the comparison baselines.

The second question \textbf{Q2} investigates the life cycle of the label arrival (Section \ref{subsec2.3}) in the stream and to which of the patterns the method is engineered to deal with.

The third question \textbf{Q3} covers whether a concept drift (Section \ref{drift}) detection mechanism is built on, how it is implemented, and what patterns are recognized.

The fourth question \textbf{Q4} examines if there is a mechanism for concept evolution (Section \ref{evol}) detection and, if yes, how it is implemented, if it detects one or more classes, and if recurring classes are considered.

The fifth question \textbf{Q5} is meant to analyze both what and how the evaluation methods are employed. The first inquiry is answered with the union of a subset of the metrics described in Table \ref{tab:02} in Section \ref{ev1} and possibly a set of metrics not described. The latter is answered with one of the procedures described in Section~\ref{ev2}.

Finally, the sixth question \textbf{Q6} is concerned with the author's view of his own work and the prospects for the future given what has been obtained so far.

\subsection{Search Criteria}\label{searchc}

In order to retrieve the most relevant work done and published in the field of interest, the search string plays a major role. As mentioned in Section \ref{meth}, PICOC criteria are components to define a plain question for a systematic review \cite{Petticrew2006}. These criteria were defined as listed:

\begin{enumerate}
\item Population: specific populace groups, such as application areas and organizations. In this study, the population is the data stream classification research area;
\item Intervention: procedures, methodologies, or tools for a given problem. Here, it is the Multi-label Classification domain;
\item Comparison: defines how to compare the intervention. This criterion is not applicable to our~case;
\item Outcome: these are the expected results. So, the outcome is an overview of algorithms for multi-label data stream classification, concept evolution detection, and concept drift detection;
\item Context: it is the study context. It involves a comparison scenario, such as the study participants and tasks performed. Our context is primary studies.
\end{enumerate}

The search strings are shown in Table~\ref{tab:04}, and were used to query five relevant databases in the computer science context, as appointed in Table~\ref{tab:05}. The search was limited to title, abstract, and keywords metadata when the database search engine allowed it.

\begin{table}[ht]
\caption{Search strings}\label{tab:04}%
\begin{tabular}{@{}llll@{}}
\toprule
Terms & Search string & Boolean operation\\
\midrule
1 & "Data stream*" & AND\\
2 & ("class*" OR "novel*" OR "detect*" OR "mining" OR "mine") & AND\\
3 & ("Multi-label" OR "Multilabel" OR "Multi-target" OR  "Multitarget")	 \\
\bottomrule
\end{tabular}
\end{table}

\begin{table}[ht]
\caption{Data Sources}\label{tab:05}%
\begin{tabular}{@{}llll@{}}
\toprule
Data source & URL\\
\midrule
ACM Digital Library & http://portal.acm.org \\
IEEE Xplore & http://ieeexplore.ieee.org \\
Science Direct & http://www.sciencedirect.com \\
Springer Link & http://link.springer.com \\
Scopus & http://www.scopus.com \\
\bottomrule
\end{tabular}
\end{table}

\subsection{Inclusion and Exclusion Criteria}\label{inc}

The period considered was January 2016 to December 2024, when the query was performed. Given the recency of the subject, papers no older than 2016 are taken as mandatory to be reviewed. Besides, the subject's highly universal nature makes it essential to be published in English, so we only considered papers written in such a language. The pertinence to the theme, as well as the primary nature of the study, was considered in the inclusion decision. Finally, the total availability of the paper was mandatory. Table~\ref{tab:06} sums up all the criteria.

\begin{table}[ht]
\caption{Inclusion and Exclusion Criteria}\label{tab:06}%
\begin{tabular}{@{}llll@{}}
\toprule
 & \textbf{Inclusion criteria}\\
IC1 & Papers in English; \\
IC2 & Recent studies (published from 2016 onwards); \\
IC3 & Studies published and fully available in selected search sources; \\
IC4 & Studies related to Data Streams and Multi-label Classification. \\
 & \textbf{Exclusion criteria}\\
EC1 & Duplicated studies; \\
EC2 & \makecell[l]{Secondary studies, short papers (four pages or less), books, technical reports,\\and other forms of gray literature;} \\
EC3 & Studies published before 2016; \\
EC4 & Studies that are not in English; \\
EC5 & Studies that do not address Data Streams and Multi-label Classification; \\
EC6 & Studies that were not available in the selected research sources. \\
\bottomrule
\end{tabular}
\end{table}

\subsection{Quality Assessment}\label{qual}


A natural metric to assess the relevance of a work is the number of citations. However, newer studies that may also be relevant can be prejudiced by such a criterion. So, in order to make justice, a simple score to balance its amount of citations and its age is proposed, such that the relevance is retrieved by the binarization of such score, according to the following system:

\begin{equation}
Is~it~relevant? = \begin{cases}
yes, & \mbox{if } \mbox{ $((1 - (2024 - year)/10)^{1.2} \times citations) \geq 3.0$} \\
no, & \mbox{otherwise}
\end{cases}
\end{equation}

This threshold of 3, in the extreme dates, corresponds to a paper from 2016 cited a minimum of 21 times or a paper from 2024 cited a minimum of 3 times.

However, relevance was not the sole quality criterion considered, as papers that did not get much attention are not necessarily of poor quality. As such, we also assessed the \textit{plainness of the explanation of the method}, the \textit{assertiveness of the purpose stated} (that is, if it is a new classification method, a novelty detection method, an application study, and so forth) and the robustness of the evaluation method (if it is evaluated through enough metrics and compares to pertinent baselines). The quality assessment checklist is given in Table~\ref{tab:07}. Answering \textit{Yes} adds up one point, as answering \textit{No} adds up nothing. The cutoff score to be an accepted paper was 3.

\begin{table}[htbp]
    \caption{Quality assessment checklist}\label{tab:07}
    \centering
    \begin{tabular}{c|c:c}
    \toprule
        Question & Answers & Score\\
    \midrule   
        \multirow{2}{*}{Is it relevant?}&Yes&1\\
        &No&0\\
        \hline
        \multirow{2}{*}{Is the purpose clearly stated?}&Yes&1\\
        &No&0\\
        \hline
        \multirow{2}{*}{Is the method well detailed?}&Yes&1\\
        &No&0\\
        \hline
        \multirow{2}{*}{Is the evaluation method robust?}&Yes&1\\
        &No&0\\
        \hdashline
        \\[-1em]
        \multicolumn{3}{r}{\makecell*[{{p{0.85cm}}}]{Max:4\\Cutoff:3}}\\
    \bottomrule
    \end{tabular}
\end{table}

\subsection{Study selection}


Selecting the studies consisted of five steps. In step 1, we queried the search string in the data sources, as specified in Section \ref{searchc}. In Step 2, we removed the duplicate studies and checked the remaining works' titles, abstracts, and keywords to check for the inclusion and exclusion criteria, as in Section \ref{inc}. In Step 3, we integrally read the previously selected studies, again checking for the inclusion and exclusion criteria. In Step 4, we performed the \textit{Snowballing} by checking the references of the works selected in Step 3 and adding them manually, if coherent with the criteria. Finally, in Step 5, we analyzed all the gathered studies with the quality assessment checklist, as described in Section \ref{qual}, removing the studies that did not score minimally three points.

\begin{figure}[hbtp]
\centering
\includegraphics[scale=0.272]{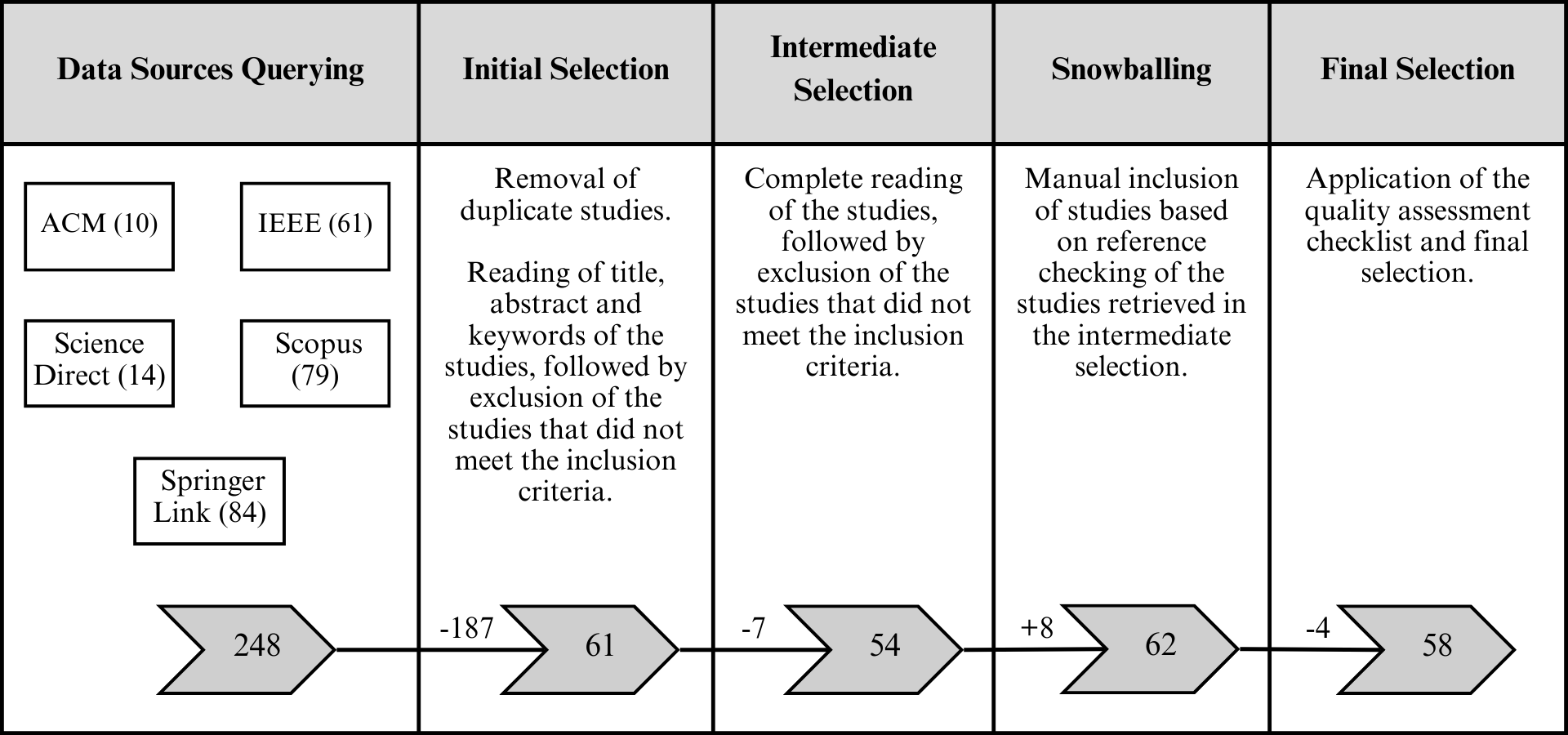}
    \caption{Study selection flow of the present Systematic Literature Review.}
\label{fig:3}
\end{figure}

In Figure \ref{fig:3}, the whole process is depicted. Initially, we retrieved 192 studies. After all the steps, 58 studies were selected, listed in the Table \ref{tab:08}.

\begin{longtable}{p{0.3cm}p{11.5cm}p{0.7cm}}
\caption{SLR selected studies}\label{tab:08}%
\\\hline
1 & A novel progressive multi-label classifier for class-incremental data~\cite{Dave2016} & 2016\\
2 & An online universal classifier for binary, multi-class and multi-label classification~\cite{MengJooEr2016} & 2016\\
3 & Online multi-label classification with adaptive model rules~\cite{Sousa2016} & 2016\\
4 & A novel online multi-label classifier for high-speed streaming data applications~\cite{Venkatesan2017} & 2017\\
5 & Label Powerset for Multi-label Data Streams Classification with Concept Drift~\cite{Costa2017} & 2017\\
6 & Multi-label classification via multi-target regression on data streams~\cite{Osojnik2016} & 2017\\
7 & Weighted ensemble classification of multi-label data streams~\cite{Wang2017} & 2017\\
8 & A label compression method for online multi-label classification~\cite{Ahmadi2018} & 2018\\
9 & A Novel Online Stacked Ensemble for Multi-Label Stream Classification~\cite{Bykakir2018} & 2018\\
10 & Multi-label classification from high-speed data streams with adaptive model rules and random rules~\cite{Sousa2018} & 2018\\
11 & Multi-label kNN Classifier with Self Adjusting Memory for Drifting Data Streams~\cite{Roseberry2018} & 2018\\
12 & Multi-Label Learning with Emerging New Labels~\cite{Zhu2018} & 2018\\
13 & Online multi-label dependency topic models for text classification~\cite{Burkhardt2018} & 2018\\
14 & Social stream classification with emerging new labels~\cite{Mu2018} & 2018\\
15 & An Online Variational Inference and Ensemble Based Multi-label Classifier for Data Streams~\cite{Nguyen2019} & 2019\\
16 & Co-training Based on Semi-Supervised Ensemble Classification Approach for Multi-label Data Stream~\cite{Chu2019} & 2019\\
17 & Dynamic Multi-label Learning with Multiple New Labels~\cite{Wang2019} & 2019\\
18 & Dynamic principal projection for cost-sensitive online multi-label classification~\cite{MinChu2019} & 2019\\
19 & Efficient ensemble classification for multi-label data streams with concept drift~\cite{Sun2019} & 2019\\
20 & Modeling Multi-label Recurrence in Data Streams~\cite{Ahmadi2019} & 2019\\
21 & Multi-label classification via incremental clustering on an evolving data stream~\cite{Nguyen2019} & 2019\\
22 & Multi-label classification via label correlation and first order feature dependance in a data stream~\cite{Nguyen2019_label} & 2019\\
23 & Multi-Label Punitive KNN with Self-Adjusting Memory for Drifting Data Streams~\cite{Roseberry2019} & 2019\\
24 & Novelty Detection for Multi-Label Stream Classification~\cite{CostaJunior2019-BR} & 2019\\
25 & Pruned Sets for Multi-Label Stream Classification without True Labels~\cite{CostaJunior2019-PS} & 2019\\
26 & An Incremental Kernel Extreme Learning Machine for Multi-Label Learning With Emerging New Labels~\cite{Kongsorot2020} & 2020\\
27 & Imbalanced Continual Learning with Partitioning Reservoir Sampling~\cite{Kim2020} & 2020\\
28 & Online Metric Learning for Multi-Label Classification~\cite{Gong2020} & 2020\\
29 & Online Semi-supervised Multi-label Classification with Label Compression and Local Smooth Regression~\cite{Li2020} & 2020\\
30 & Robust Online Multilabel Learning Under Dynamic Changes in Data Distribution With Labels~\cite{Du2020} & 2020\\
31 & A new self-organizing map based algorithm for multi-label stream classification~\cite{Cerri2021} & 2021\\
32 & An Efficient Framework for Multi-Label Learning in Non-stationary Data Stream~\cite{Zheng2021} & 2021\\
33 & Multi-Label kNN classifier with Online Dual Memory on data stream~\cite{Wang2021} & 2021\\
34 & Probabilistic Label Tree for Streaming Multi-Label Learning~\cite{Wei2021} & 2021\\
35 & Self-adjusting k nearest neighbors for continual learning from multi-label drifting data streams~\cite{Roseberry2021} & 2021\\
36 & Adaptive ensemble of self-adjusting nearest neighbor subspaces for multi-label drifting data streams~\cite{Alberghini2022} & 2022\\
37 & An Algorithm Adaptation Method for Multi-Label Stream Classification using Self-Organizing Maps~\cite{Cerri2022} & 2022\\
38 & Evolving multi-label fuzzy classifier~\cite{Lughofer2022} & 2022\\
39 & High-Dimensional Multi-Label Data Stream Classification With Concept Drifting Detection~\cite{Li2022} & 2022\\
40 & Improved Ensemble Classification for Evolving Data Streams~\cite{Tian2022} & 2022\\
41 & Incremental deep forest for multi-label data streams learning~\cite{Liang2022} & 2022\\
42 & ML-KELM: A Kernel Extreme Learning Machine Scheme for Multi-Label Classification of Real Time Data Stream in SIoT~\cite{Luo2022} & 2022\\
43 & ROSE: robust online self-adjusting ensemble for continual learning on imbalanced drifting data streams~\cite{Cano2022} & 2022\\
44 & Semi-Supervised Online Kernel Extreme Learning Machine for Multi-Label Data Stream Classification~\cite{Qiu2022} & 2022\\
45 & A Weighted Ensemble Classification Algorithm Based on Nearest Neighbors for Multi-Label Data Stream~\cite{Wu2023} & 2023\\
46 & Aging and Rejuvenating Strategies for Fading Windows in Multi-Label Classification on Data Streams~\cite{Roseberry2023} & 2023\\
47 & Methods for Predicting the Rise of the New Labels from a High-Dimensional Data Stream~\cite{Kanagaraj2023} & 2023\\
48 & Novelty detection for multi-label stream classification under extreme verification latency~\cite{Costa2023} & 2023\\
49 & Online Label Distribution Learning Using Random Vector Functional-Link Network~\cite{Huang2023} & 2023\\
50 & Online Multi-Label Classification with Scalable Margin Robust to Missing Labels~\cite{Zou2023} & 2023\\
51 & Online Semi-Supervised Classification on Multilabel Evolving High-Dimensional Text Streams~\cite{Kumar2023} & 2023\\
52 & Unsupervised concept drift detection for multi-label data streams~\cite{Gulcan2023} & 2023\\
53 & Application Research of Multi-label Learning Under Concept Drift~\cite{Tang2024} & 2024\\
54 & Balancing efficiency vs. effectiveness and providing missing label robustness in multi-label stream classification~\cite{Bakhshi2024} & 2024\\
55 & Hoeffding adaptive trees for multi-label classification on data streams~\cite{Esteban2024} & 2024\\
56 & Imbalance-Robust Multi-Label Self-Adjusting kNN~\cite{Nicola2024} & 2024\\
57 & Prioritized Binary Transformation Method for Efficient Multi-label Classification of Data Streams with Many Labels~\cite{Yildirim2024} & 2024\\
58 & Weak Multi-Label Data Stream Classification Under Distribution Changes in Labels~\cite{Zou2024}
& 2024\\
\bottomrule
\end{longtable}

\subsection{Data Extraction}\label{dex}

The final selected papers underwent a data extraction process. The composition of the data extraction form was guided by the research questions defined in Section \ref{rese}. Table \ref{tab:09} shows the data extraction form.

\begin{longtable}{p{3cm}p{5cm}p{3cm}}
\caption{Extracted data from the SLR studies}\label{tab:09}%
\\\hline
\textbf{Field} & \makecell{\textbf{Possible values}} & \makecell[r]{\textbf{Research question}}\\
\midrule
Authors &  & \makecell[r]{Overview}\\
\hline
Publication year & \makecell{\{2016, \dots, 2024\}} & \makecell[r]{Overview}\\
\hline
Method name & & \makecell[r]{Q1}\\
\hline
\makecell[l]{Approach} & \makecell{Algorithm adaptation\\Problem transformation\\Ensemble} & \makecell[r]{Q1}\\
\hline
\makecell[l]{Label latency} & \makecell{Infinite\\Delayed\\Immediate}& \makecell[r]{Q2}\\
\hline
\makecell[l]{Patterns of drift detected} & \makecell{Abrupt\\Gradual\\Incremental\\Recurring\\Does not detect\\Not mentioned
} & \makecell[r]{Q3}\\
\hline
\makecell[l]{How is concept evolution\\approached?} & \makecell{Concept evolution\\detection\\Recurring and concept\\evolution detection\\
Not approached} & \makecell[r]{Q4}\\
\hline
\makecell[l]{How many new classes are\\simultaneously detected?} & \makecell{1\\2+\\None} & \makecell[r]{Q4}\\
\hline
Dataset used & & \makecell[r]{Overview}\\
\hline
\makecell[l]{How are the evaluation\\methods employed?} & \makecell{Prequential\\Holdout\\Not mentioned} & \makecell[r]{Q5}\\
\hline
\makecell[l]{Is the data available?} & \makecell{Yes\\Partially\\Not mentioned} & \makecell[r]{Overview}\\
\hline
\makecell[l]{Is the code available} & \makecell{Yes\\Not mentioned} & \makecell[r]{Overview}\\
\hline
\makecell*[{{p{3.5cm}}}]{Comparison baselines} & & \makecell[r]{Q1}\\
\hline
\makecell*[{{p{3.5cm}}}]{Information about memory and runtime} & & \makecell[r]{Overview}\\
\hline
\makecell*[{{p{3.5cm}}}]{Information about asymptotic complexity} & & \makecell[r]{Overview}\\
\hline
\makecell[l]{Does it use pre-built\\libraries?} & \makecell{MOA\\MEKA\\Scikit-multiflow / River\\Apache Kafka\\Not mentioned} & \makecell[r]{Overview}\\
\hline
\makecell*[{{p{3.5cm}}}]{Future work proposals} & & \makecell[r]{Q6}\\
\hline
\end{longtable}

\section{Exploratory analysis}\label{sec:exploratory}

This section concerns the answers to the research questions defined in Section \ref{rese}, given according to the findings made possible by the data extraction form in Section \ref{dex}.

Before diving into the questions, we performed an exploratory analysis of the gathered data, aiming to give an overview of what has been produced in the field over the last years.

\subsection{Co-occurence}

\begin{figure}[hbtp]
\centering
\includegraphics[scale=0.13]{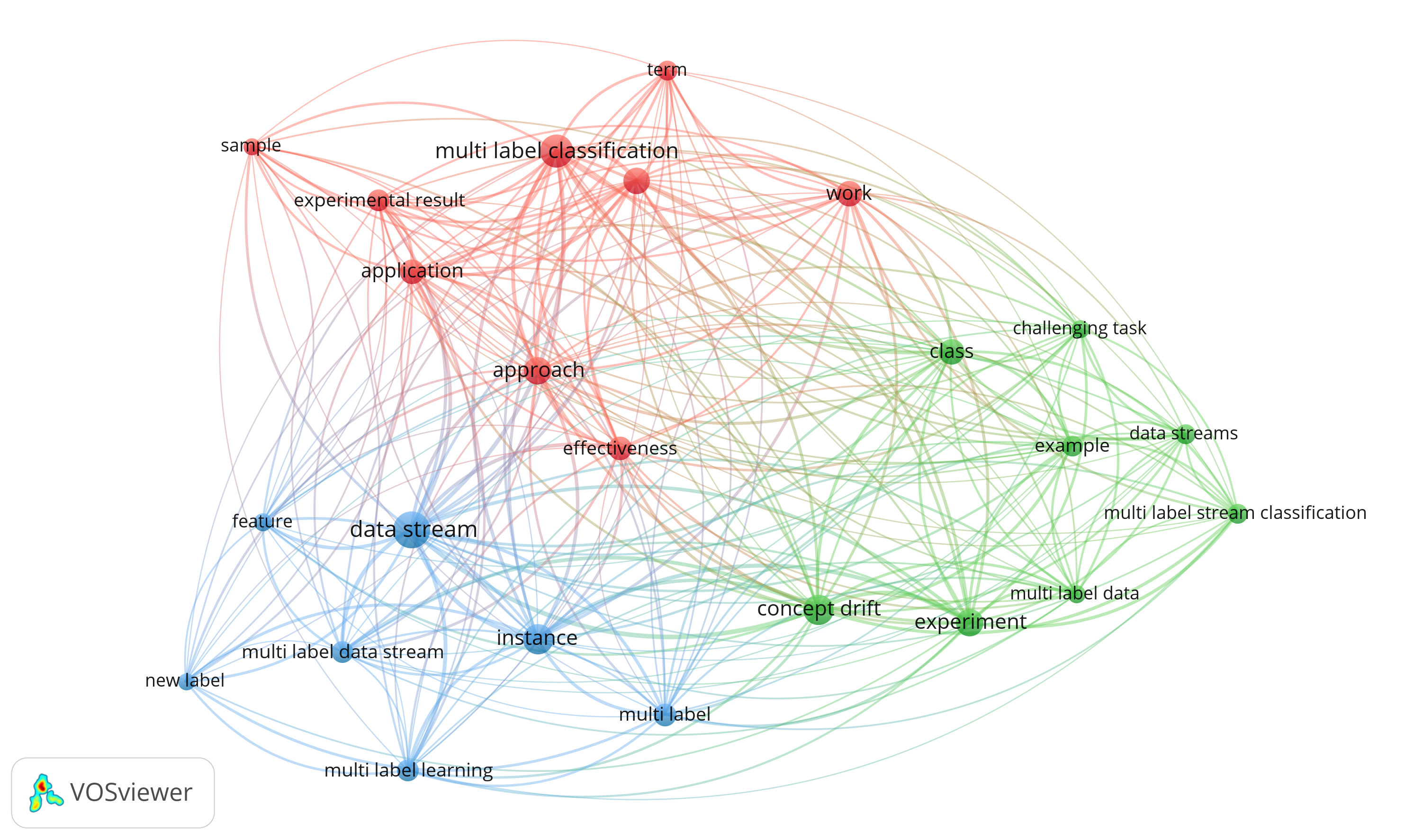}
    \caption{Co-occurrence of concepts in title and abstract.}
\label{fig:4}
\end{figure}

A co-occurrence map of the concepts within the extracted papers can be seen in Figure~\ref{fig:4}. The map shows the 60\% most relevant words in the abstract or title of at least seven publications. It can be clearly seen that the VOSviewer tool identified three clusters that ultimately revolve around the concepts ``data stream'', ``concept drift'', and ``instance''.

The figure shows us the presence of 3 clusters: one revolving ``data stream'', another revolving ``multi-label classification'', and the third revolving ``concept drift''.

Even though the first two centers are expected, ``concept drift'' as a cluster center is surprising, even more so considering that neither ``concept'' nor ``drift'' were part of the search string. This fact shows us how closely these concepts are: it is not easy to achieve good results in data stream learning without considering the change in data distribution. Worthy of note is the presence of the concept ``new label'', which shows up with a small weight, being an immediate reflection of how much more attention concept drift has gained in spite of concept evolution. Also, the same item does not connect to the ``concept drift'' item, which is also a reflection that these problems tend to be addressed independently, with rare exceptions, as will be seen in our research questions.

\begin{figure}[hbtp]
\centering
\includegraphics[scale=0.13]{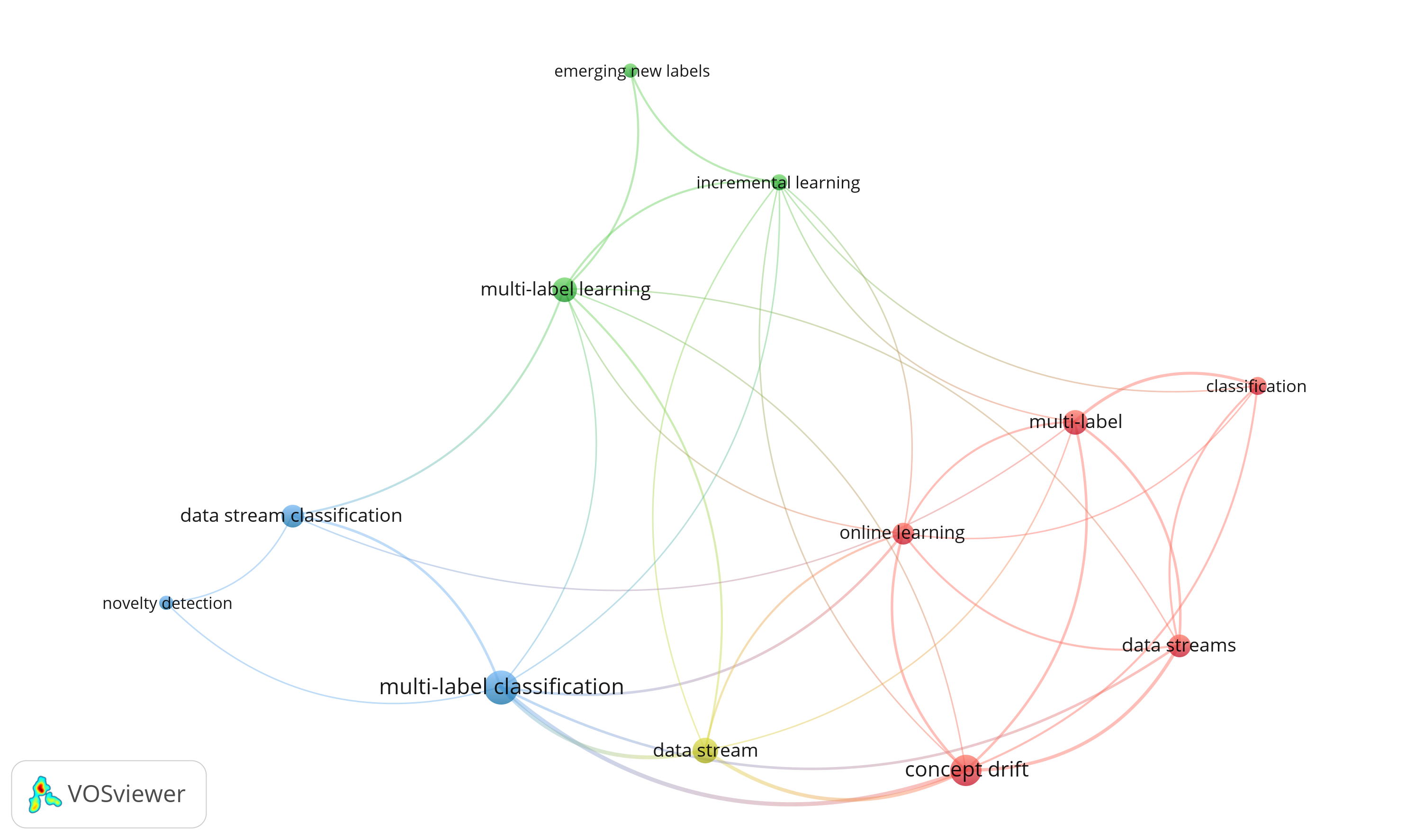}
    \caption{Co-occurrence of keywords.}
\label{fig:5}
\end{figure}

Figure~\ref{fig:5} is also a co-occurrence map but concerns the keywords that appear in at least three publications. Once again, ``concept drift'' comes as a relevant token, being as relevant as the subject itself, that is, ``multi-label classification''. Even though ``concept evolution'', an important theme addressed in this review, does not appear explicitly, the keywords ``novelty detection'', as defined, should embrace the idea of concept evolution. Yet, the keyword ``emerging new labels'' stands exactly for the same as concept evolution. It is clear that it is a far less addressed issue than concept~drift.

\subsection{Publications}
\begin{figure}[hbtp]
\centering
\includegraphics[scale=0.6]{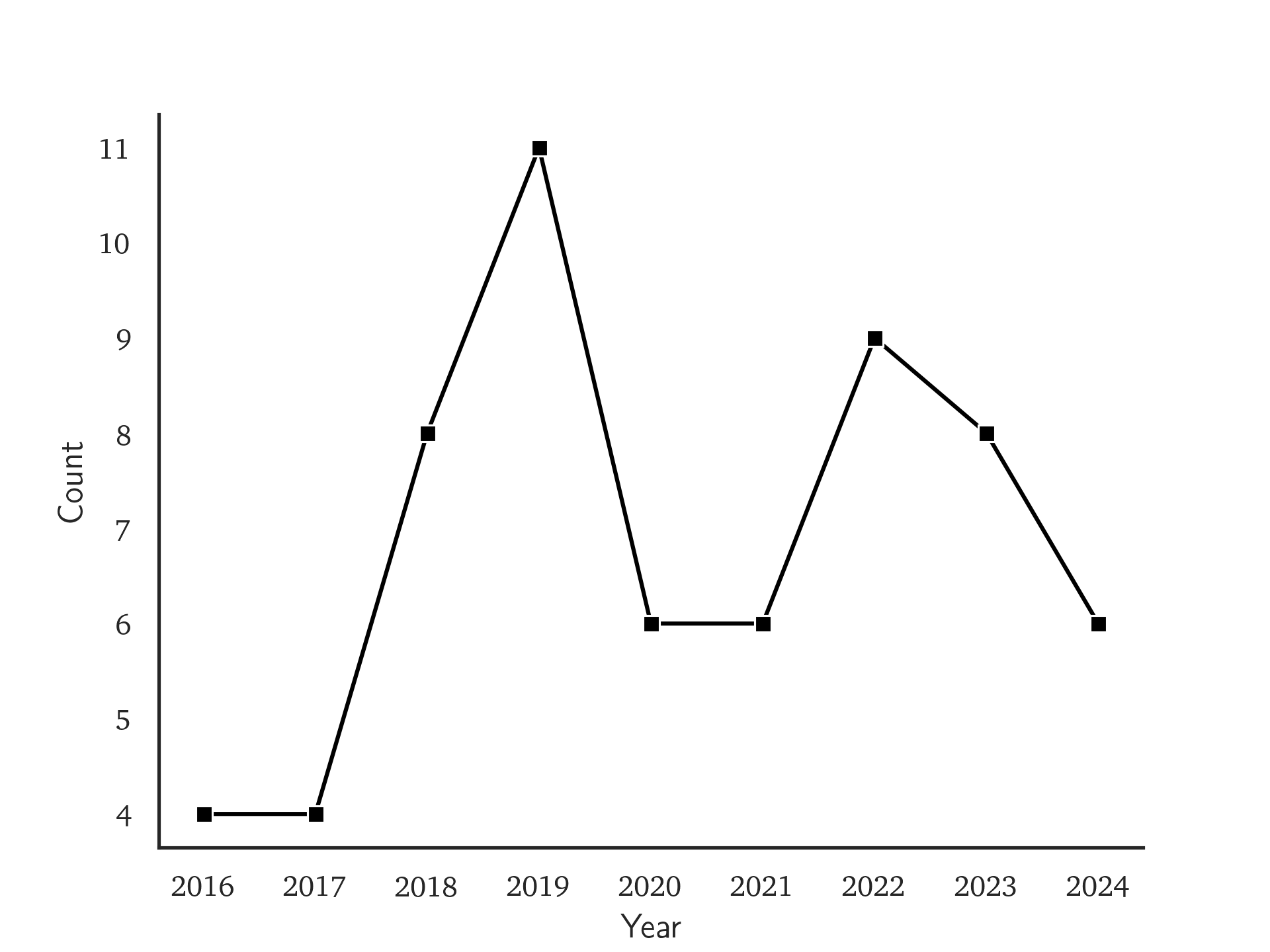}
    \caption{Number of publications per year.}
\label{fig:6}
\end{figure}

Figure~\ref{fig:6} depicts the number of publications per year in the period analyzed. The period begins with 2 years of little production, rising in 2028, and climaxing at a significant peak in 2019. Then, we have a decrease, but that still does not fall to the low levels of 2016 and 2017, which seems to point to a resurgence of the interest in the data streams topic that has been somewhat constant ever since, characterized by a mean of 7-8 publications per year (since it started to rise again in 2018).

\subsection{Datasets} 
\begin{figure}[hbtp]
\centering
\includegraphics[scale=0.8]{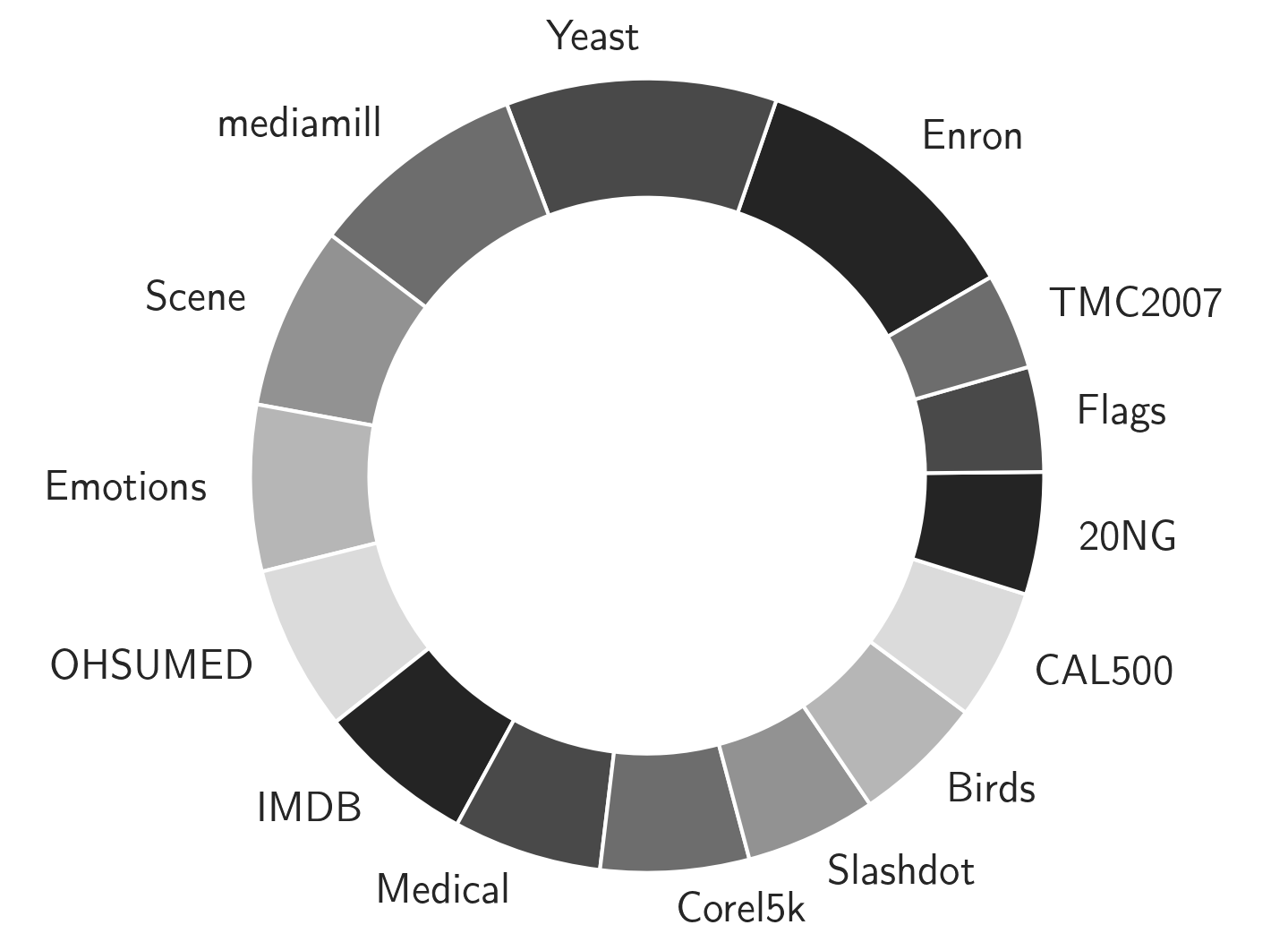}
    \caption{Top 15 most used datasets.}
\label{fig:7}
\end{figure}

Concerning the datasets used, the most used ones are classical \textit{batch} multi-label classification datasets, which are then streamed by MOA or other tools. Figure~\ref{fig:7} depicts the datasets that are the most used in studies.

Some papers, mainly aiming at testing their concept drift mechanisms, make use of synthetic datasets. \citet{CostaJunior2019-PS, CostaJunior2019-BR} use an RBF (Radial Basis Function) generator, available in MOA. \citet{Costa2017, Wang2021} use both RBF and RTG (random tree generator). \citet{Nguyen2019} not only generate a synthetic RTG dataset, but also introduce drift in classical \textit{batch} multi-label datasets by concatenating parts of the dataset with parts generated by RTB. \citet{Alberghini2022} make use of a hypersphere generator. \citet{Cerri2021, Cerri2022} use RBF, waveform, RTG, and hyper-plane generators. \citet{Kumar2023} use the TextGenerator present in MOA. \citet{Esteban2024} use RTG, RBF, and hyper-plane generators. \citet{Bakhshi2024} use SEA. \citet{Cano2022} use Agrawal, AssetNegotiation, RandomRBF, SEA, Sine, and Hyper-plane. Finally, \citet{Ahmadi2019} use a generator built from scratch that simulates recurrent concepts.

\subsection{Libraries}
\begin{figure}[hbtp]
\centering
\includegraphics[scale=0.8]{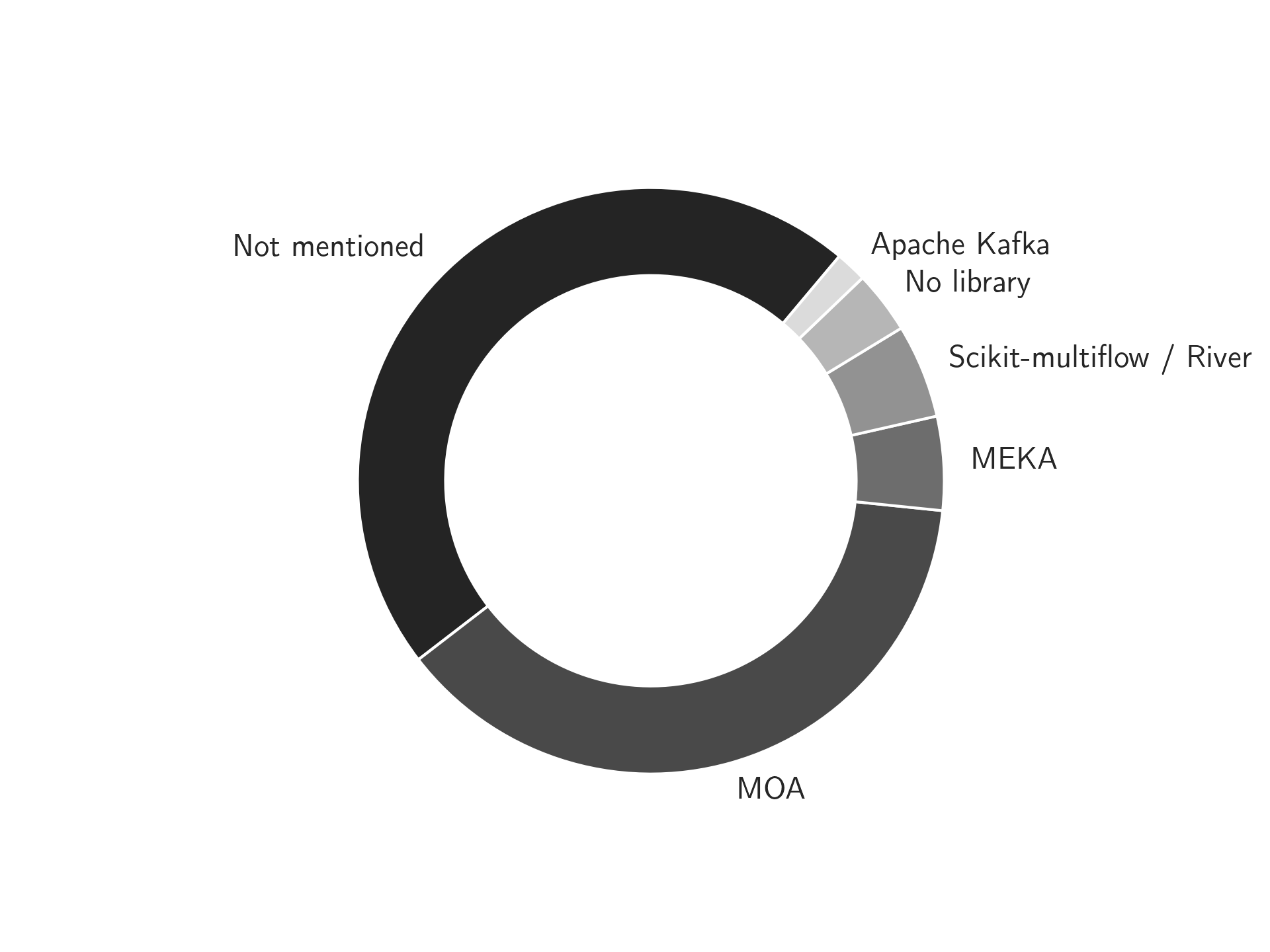}
    \caption{Pre-built libraries usage.}
\label{fig:8}
\end{figure}

There are several ways to handle streaming data. It is usual to use pre-built libraries to make it easier to build and test multi-label classifiers in this setting. By far, the most used is MOA (as already cited in the \hyperref[sec1]{Introduction}). Second comes MEKA~\cite{Read2016}, which is actually meant for offline learning and was used as a source of benchmarking batch classifiers. Then, Scikit-multiflow (a MOA ``equivalent'' in Python) and its successor, River. Lastly, Apache Kafka~\cite{Kreps2011} simulates a data stream at a fixed speed. Figure~\ref{fig:8} depicts the usage of those pre-built libraries in the reviewed studies.

\subsection{Time and space complexity}
From all papers we reviewed, 27 are explicit about their asymptotic complexity, at least concerning the time complexity. Table~\ref{tab:10} brings them all together.

\renewcommand\theadfont{\bfseries}

\begin{longtable}{|p{1.9cm}|p{3.6cm}|p{2.3cm}|p{4.8cm}|}
\caption{Asymptotic Complexities of Various Methods}\label{tab:10}\\
\hline
 & \multicolumn{2}{l|}{Asymptotic complexity} &  \\ \hline
\endfirsthead

\multicolumn{4}{c}%
{{\bfseries \tablename\ \thetable{} -- continued from previous page}} \\
\hline
 & \multicolumn{2}{l|}{Asymptotic complexity} &  \\ \hline
\endhead

\hline \multicolumn{4}{r}{{Continued on next page}} \\
\endfoot

\hline
\endlastfoot

Method & \multicolumn{1}{l|}{Time} & Space & Notes \\ \hline
ML-AMRules & $O(n)$ &  & \textbf{n} = input size \\ \hline
GOOWE-ML & $O(NK^2L)$ &  & \makecell[l]{\textbf{N} = number of instances\\ \textbf{K} = ensemble size\\ \textbf{L} = number of labels} \\ \hline
MuENL / MuENLHD & \makecell[l]{$O(|v|^2nd')$\\ $O(|v|^2nd')$\\ $O(|q|ge_m/\psi)$\\ $O(mdn)$\\ $O(d'mn)$} &  & \makecell[l]{\textbf{|v|} = size of class label set\\ \textbf{n} = number of instances\\ \textbf{d} = number of dimensions \\ \textbf{d'} = number of features at the \\output of FDUpdate\\ \textbf{m} = number of random features \\ \textbf{$\psi$} = number of instances having \\tree height limit $e_m$} \\ \hline
NL-Forest & \makecell[l]{Training: $O(Z\psi\log\psi +$\\ $ cz\phi\log\phi)$\\ Prediction: $O(Z\log\psi +$\\ $ uz\log\phi)$\\ Update: $O(Zs\log s)$ or\\ $O(z\phi\log\phi)$} & $O(s + Zd\psi)$ & \makecell[l]{\textbf{Z} = number of trees in the forest\\ \textbf{$\psi$} = sample size for I-F\\ \textbf{c} = number of subforests\\ \textbf{z} = number of trees in each\\subforest\\ \textbf{$\phi$} = sample size for L-F\\ \textbf{u} = number of main labels\\ \textbf{s} = buffer size} \\ \hline
SCVB-Dependency & Fast-Dep.-LLDA: $O(jt + kd)$ &  & \makecell[l]{\textbf{jt} =number of labels in topic t\\\textbf{kd} = number of topics in\\document d} \\ \hline
MLSAMPkNN & $O(m_{max} \log_2 (\frac{m_{max}}{m_{min}})$ &  & \textbf{m} = size of the window \\ \hline
CS-DPP & \makecell[l]{DPP-PBC: $O(d^2 + MK +$\\$ Kd + M^2K)$\\ DPP-PBT: $O(d^2 + M^2d +$\\$M^2K)$} & \makecell[l]{DPP-PBC: $O(d^2 + $\\$MK + Kd)$\\ DPP-PBT: $O(d^2 +$\\$ MK + Md)$} & \makecell[l]{\textbf{d} = feature vector dimension\\ \textbf{K} = number of labels\\ \textbf{M} = dimension of the code space} \\ \hline
ML-LC & $O(\max(hMD^2 + hD, |Y|^2 + |Y|D^2))$ & $O(M(Dn^2 + D^2))$ & \makecell[l]{\textbf{h} = number of predicted labels\\ \textbf{M} = number of labels\\ \textbf{D} = number of features\\ \textbf{|Y|} = cardinality of the true\\label set\\ \textbf{n} = average number of values for\\features} \\ \hline
ML-IK & Matrix inversion: $O(N^3)$ or $O(N^{2.807})$ &  & \textbf{N} = size of kernel \\ \hline
ELM-OMLL &  &  &  \\ \hline
OnSeML & $O(nd^3 + (\frac{n}{sQ}) \times \min(k^3, s^3Q))$ &  & \makecell[l]{\textbf{n} = number of instances\\ \textbf{d} = number of features\\ \textbf{sQ} = number of labeled instances\\to update\\ \textbf{k} = dimension of the encoded\\label set} \\ \hline
OML & \makecell[l]{Training: $O(np + dpq)$\\ Testing: $O(dpq + nd)$} &  & \makecell[l]{\textbf{n} = number of instances\\ \textbf{p} = number of features\\ \textbf{q} = number of labels\\ \textbf{d} = dimension of the new\\projection space} \\ \hline
ODM & $O(s+r)$ &  & \makecell[l]{\textbf{s} = pairs of reservoir and\\associated prototypes\\ \textbf{r} = size of reservoirs} \\ \hline
MLSAkNN & $O(m_{max}d + m_{max}L + m_{max}\log_2(\frac{m_{max}}{m_{min}})$ & $O((d + L)m_{max} + m_{max}^2)$ & \makecell[l]{\textbf{m} = current window size\\between $m_{min}$ e $m_{max}$.\\ \textbf{d} = dimensionality of an instance\\ \textbf{L} = number of possible labels} \\ \hline
PSLT & $O(m(dK^D + dK\Delta y \sim y\epsilon p + N^D) + |L_T| \tau_tK^D)$ & $O((K + 1) (|V_T | - |L_T|) D + |L_T|^D)$ & \makecell[l]{\textbf{$\Delta y \sim y$} = average discrepancy\\between the meta-label vector y\\and the modified label vector $\sim$y\\ \textbf{m} =number of emerging labels\\ \textbf{K} = branch factor for tree node\\ \textbf{D} = number of non-zero features\\of node representations of average\\ \textbf{$\epsilon$} = solving precision\\ \textbf{p} =?\\ |L\textsubscript{T}| = size of the set of leaf nodes\\of T\\ |V\textsubscript{T} | = size of the set of nodes\\of T} \\ \hline
ROSE & $O(2k\lambda|S|)$ & $O((2krvlc) + (|w|f))$ & \makecell[l]{\textbf{k} = number of base classifiers in\\the ensemble\\ \textbf{$\lambda$} = Poisson $\lambda$ for online bagging\\ \textbf{|S|} = total size of the data flow\\ \textbf{r} = dimension of the feature\\subspace\\ \textbf{v} = maximum number of values\\per feature\\ \textbf{l} = number of leaves in the tree\\ \textbf{c} = number of classes\\ \textbf{|w|} = sliding window buffer size\\ \textbf{f} = number of features} \\ \hline
VDSDF & $O(TWn^3)$ & $O(n^3)$ & \makecell[l]{\textbf{T} = maximum number of layers\\of the model\\ \textbf{W} =number of basic units in\\each layer\\ $n^3$ = complexity of previous\\algorithms 1 and 2} \\ \hline
ML-KELM & \makecell[l]{Matrix inversion: $O(n^3)$\\ E-IL update: $O(k \times m(t)^2)$\\ C-IL update: $O(k^3)$} &  & \makecell[l]{\textbf{n} = number of instances\\ \textbf{k} = data block size in E-IL or\\training set size for C-IL\\ \textbf{m(t)} = number of trained\\instances until time t} \\ \hline
ML-MRMR-FS & $O(nMLK)$ &  & \makecell[l]{\textbf{n} = number of instances in a\\data chunk\\ M = number of features\\ L = number of possible labels\\ K = size of the ensemble model} \\ \hline
OSMTS & $O(\bar{K}(\bar{d}\bar{V})+\bar{K}\log(\bar{K}))$ & $O(\bar{K}(3\bar{V} + \bar{V}^2) + VD)$ & \makecell[l]{$\bar{K}$ = average number of\\micro-clusters\\ $\bar{V}$ = average number of words\\in the feature set of each cluster\\ $\bar{d}$ =  average size of arriving\\document\\ \textbf{V} = size of the active vocabulary\\ \textbf{D} = number of active documents} \\ \hline
ARkNN & $O(mkd + |L|)$ &  & \makecell[l]{\textbf{m} = maximum size of the window\\ \textbf{k} = number of neighbors\\ \textbf{d} = dimensionality of data\\ \textbf{|L|} = number of labels} \\ \hline
LD3 & $O(wn^2)$ &  & \makecell[l]{\textbf{w} = number of samples inside\\a time window\\ \textbf{n} = number of labels} \\ \hline
WENNML & $O(h1 \cdot I) + O(h2 \cdot I)$ &  & \makecell[l]{\textbf{I} = number of base classifiers for\\both AEC and PEC\\ \textbf{h1} = number of instances in the\\fixed size data block for AEC\\ \textbf{h2} = number of instances in the\\variable size data block for PEC} \\ \hline
OML-SM & $O(T (dq + q^2))$ & $O(dq + q^2)$ & \makecell[l]{\textbf{T} = total number of samples\\ \textbf{d} = feature dimensionality of\\data sets\\ \textbf{q} = number of class labels} \\ \hline
IRMLSAkNN & $O(m_{max}n + m_{max}q + m_{max}\log_2 \frac{m_{max}}{m_{min}})$ &  & \makecell[l]{\textbf{m} = window size\\ \textbf{n} = number of attributes\\ \textbf{q} = number of labels} \\ \hline
RMSC & $O(kq(T_0 + T) + q^2T_0)$ & $O(dq)$ & \makecell[l]{\textbf{T} = total number of examples\\ \textbf{T\textsubscript{0}} = size of the initialization\\chunk\\ \textbf{d} = number of features\\ \textbf{q} = number of class labels\\ \textbf{k} = average non-zero feature\\values per sample} \\ \hline
MLHAT & $O(hl^2vck)$ &  & \makecell[l]{\textbf{h} = height of the tree\\ \textbf{l} = number of labels\\ \textbf{f} = number of features\\ \textbf{v} = number of computations of\\information gains\\ \textbf{c} = number of operations for each\\information gain\\ \textbf{k} = number of base models in the\\bagging method} \\\hline
\end{longtable}

Evidently, the complexities may not be compared directly since they have different components and may not deal with the same number of problems (e.g., some are concerned with concept drift, others are not). The same can be said about Table~\ref{tab:11}, where the time and memory consumption of the methods (25 which explicitly declare them) are listed.

\begin{longtable}{|l|l|l|}
\caption{Time and memory consumption}\label{tab:11} \\
\hline
Method               & Time consumption                                                                                        & Memory consumption          \\ \hline
\endfirsthead
\multicolumn{3}{c}%
{{\bfseries \tablename\ \thetable{} -- continued from previous page}} \\
\hline
Method               & Time consumption                                                                                        & Memory consumption          \\ \hline
\endhead
\hline \multicolumn{3}{r}{{Continued on next page}}
\endfoot
\hline
\endlastfoot

ARkNN                & 0.0757 s                                                                                                 & 0.00687 RAM-H               \\ \hline
ELM-OMLL             & \begin{tabular}[c]{@{}l@{}}Training\\ Initial: 0.053\\ Sequential: 0.012\end{tabular}                   & -                           \\ \hline
GOOWE-ML             & 3481.57 s (Avg)                                                                                         & 701.76 MB                   \\ \hline
ML-BELS              & 0.0057 s / instance                                                                                    & 52.39 MB                    \\ \hline
ML-IK                & \begin{tabular}[c]{@{}l@{}}Train: 7.78 s\\ Update: 56.73 s\end{tabular}                                 & -                           \\ \hline
ML-KELM              & Train: 6217.2 s                                                                                      & -                           \\ \hline
ML-LC                & 828,71 s                                                                                                & -                           \\ \hline
ML-SAM-kNN           & Evaluation: 16394 s                                                                                     & 0.88 RAM-H                  \\ \hline
MLAW                 & 165.8 s (Avg. Total)                                                                                   & -                           \\ \hline
MLHAT                & 17.355 s (Avg)                                                                                        & -                           \\ \hline
MLSAMPkNN            & Evaluation: 0.1204 s                                                                                    & 0.4645 RAM-H                \\ \hline
MVI                  & 201.86 s (Avg. Total)                                                                            & -                       \\ \hline
NL-Forest            & \begin{tabular}[c]{@{}l@{}}Runtime per 1k items: $\sim$150 s\\ Avg. update time: $\sim$15s\end{tabular} & -                       \\ \hline
ODM                  & $\sim$158.3 s                                                                                           & -                           \\ \hline
OSML‑ELM             & \begin{tabular}[c]{@{}l@{}}Training: 1.82 s\\ Testing: 0.04 s\end{tabular}                              & -                           \\ \hline
OSMTS                & $\sim$8.208 s / instance                                                                                & $\sim$16.09 MB / instance \\ \hline
PBT                  & 0.00581 s / instance                                                                                    & 168.113 MB                  \\ \hline
RACE                 & 15128.67 s (Avg)                                                                                        & -                           \\ \hline
RMSC                 & 23.97 s (Avg Total)                                                                                     & -                           \\ \hline
ROSE                 &0.00072 s / instances                                                                               & 0.042 RAM-H                 \\ \hline
SAMEP                & \begin{tabular}[c]{@{}l@{}}Training: 1.11 s\\ Testing: 142.12 s\end{tabular}                            & -                           \\ \hline
SCVB-Dependency      & $\sim$33 s                                                                                              & -                           \\ \hline
SWMEC                & 0.0002965 s / instance                                                                                  & -                           \\ \hline
Universal Classifier & \begin{tabular}[c]{@{}l@{}}Avg. Training: 2.213s \\ Avg. Test: 0.049 s\end{tabular}                     & -                           \\ \hline
iSOUP-Tree           & 369.95 s (Avg)                                                                                          & 403.84 MB (Avg)             \\ \hline

\end{longtable}


\section{Q1: How does the classifier work?}\label{sec:q1}

\begin{figure}[htbp]
\caption{Hierarchy of methods.}
\begin{forest}
  for tree={
    align=center,
    font=\sffamily\scriptsize,
    edge+={thick},
    fork sep=3pt,
    s sep=0.08mm,
    l sep=3mm,
    grow=east,
    scale=0.94,
  },
  forked edges,
  if level=0{
    inner xsep=0pt,
    tikz={\draw [thick] (.children first) -- (.children last);}
  }{},
  [Approaches, l=1cm
    [Ensemble, l=3.8cm
        [Heterogeneous
            [GOOWE-ML~\cite{Bykakir2018}]
        ]
        [Homogeneous, l=1cm
            [Arbitrary,l=0.1cm
                [Tree
                    [ROSE~\cite{Cano2022}]
                    [MLAW~\cite{Sun2019}]
                ]
            ]
            [Defined,l=0.1cm
                [kNN
                    [AESAKNNS~\cite{Alberghini2022}$^{\Omega}$]
                    [WENNML~\cite{Wu2023}]
                    [SWMEC~\cite{Wang2017}]
                    [SWMEC$^{2}$~\cite{Tian2022}]
            ]
                [PT-SL
                    [ML-BELS~\cite{Bakhshi2024}*]
                    [Multi-label GraphPool~\cite{Ahmadi2019}$^{\zeta}$]
                ]
                [PT-Regression
                    [ML-Random Rules~\cite{Sousa2018}$^{\psi}$]
                ]
                [Probabilistic
                    [MVI~\cite{Nguyen2019_variational}]
                ]
                [Tree
                    [NL-Forest~\cite{Mu2018}]
                    [VDSDF~\cite{Liang2022}]
                ]
                [Other
                    [CMLDSE~\cite{Chu2019}, edge={red, thick}, text=red]
                ]
            ]
        ]
    ]
    [Other,l=1.3cm
        [Detection
            [LD3~\cite{Gulcan2023}]
        ]
        [Sampling
            [PRS~\cite{Kim2020}]
        ]
        [Feature Selection
        [ML-MRMR-FS~\cite{Li2022}]
        ]
    ]
    [Algorithm\\Adaptation,name=AA,l=1cm
      [kNN
        [Memory-based
            [ODM~\cite{Wang2021}]
            [MLSAkNN~\cite{Roseberry2021}$^{\Omega}$]
            [MLSAMPkNN~\cite{Roseberry2019}]
            [ML-SAM-kNN~\cite{Roseberry2018}]
            [IRMLSAkNN~\cite{Nicola2024}]
            [ARkNN~\cite{Roseberry2023}]
            [SAMEP~\cite{Zheng2021}]
        ]
        [Metric learning
            [OML~\cite{Gong2020}]
        ]
      ] 
      [Regression
        [OLD\_RVFL+*~\cite{Huang2023}]
        [EFC-ML*~\cite{Lughofer2022}]
        [OnSeML~\cite{Li2020}, edge={red, thick}, text=red]
        [CS-DPP~\cite{MinChu2019}]
      ]
      [Neural networks
        [SOM,l=2.5cm
            [SOM-AA~\cite{Cerri2022}]
        ]
        [ELM,l=4.5cm
            [Universal Classifier~\cite{MengJooEr2016}]
            [SSO-KELM~\cite{Qiu2022}, edge={red, thick}, text=red]
            [Pro-EMLC~\cite{Dave2016}]
            [OSML‑ELM~\cite{Venkatesan2017}]
            [ML-KELM~\cite{Luo2022}]
            [ML-IK~\cite{Kongsorot2020}]
            [ELM-MLCD~\cite{Tang2024}]
            [ELM-OMLL~\cite{Du2020}]
        ]
      ]
      [Probabilistic
        [SCVB-Dependency~\cite{Burkhardt2018}]
        [OSMTS~\cite{Kumar2023}, edge={red, thick}, text=red]
      ]
      [Margin-based
        [RMSC~\cite{Zou2024}]
        [OML-SM~\cite{Zou2023}]
      ]
      [Linear
        [MuENL/MuENLHD~\cite{Zhu2018}]
        [MuEMNLHD~\cite{Kanagaraj2023}]
        [MuPND~\cite{Wang2019}]
      ]
      [Trees,name=T
        [PSLT~\cite{Wei2021}]
        [MLHAT~\cite{Esteban2024}]
      ]
      [Clustering
        [OMLC~\cite{Nguyen2019}]
      ]
    ]
    [Problem\\Transformation,name=PT,l=0.5cm
        [Single label,name=SL,l=4cm
            [Binary transformation
                [PBT~\cite{Yildirim2024}]
                [MINAS-BR$^{2}$~\cite{Costa2023}]
                [MINAS-BR~\cite{CostaJunior2019-BR}]
            ]
            [Label compression
                [RACE~\cite{Ahmadi2018}\label{race}$^{\zeta}$]
            ]
            [Hybrid
                [SOM-PT~\cite{Cerri2021}]
            ]
            [Regression,name=R2
                [ML-AMRules~\cite{Sousa2016}$^{\psi}$,phantom]
            ]            
        ]        
        [Pair of labels
            [ML-LC~\cite{Nguyen2019_label}]
        ]        
        [Multiple labels,name=ML
            [Regression,name=R1,l=4.1cm
                [ML-AMRules~\cite{Sousa2016}$^{\psi}$,tier=term,forked edge,name=shared]
                [iSOUP-Tree~\cite{Osojnik2016}]
            ]
            [Powerset
                [MINAS-PS~\cite{CostaJunior2019-PS}]
                [MINAS-LP~\cite{Costa2017}]
            ]
        ]
    ]
  ]
\draw[thick] (R2) -| (shared);
\end{forest}
\label{fig:9}
\end{figure}
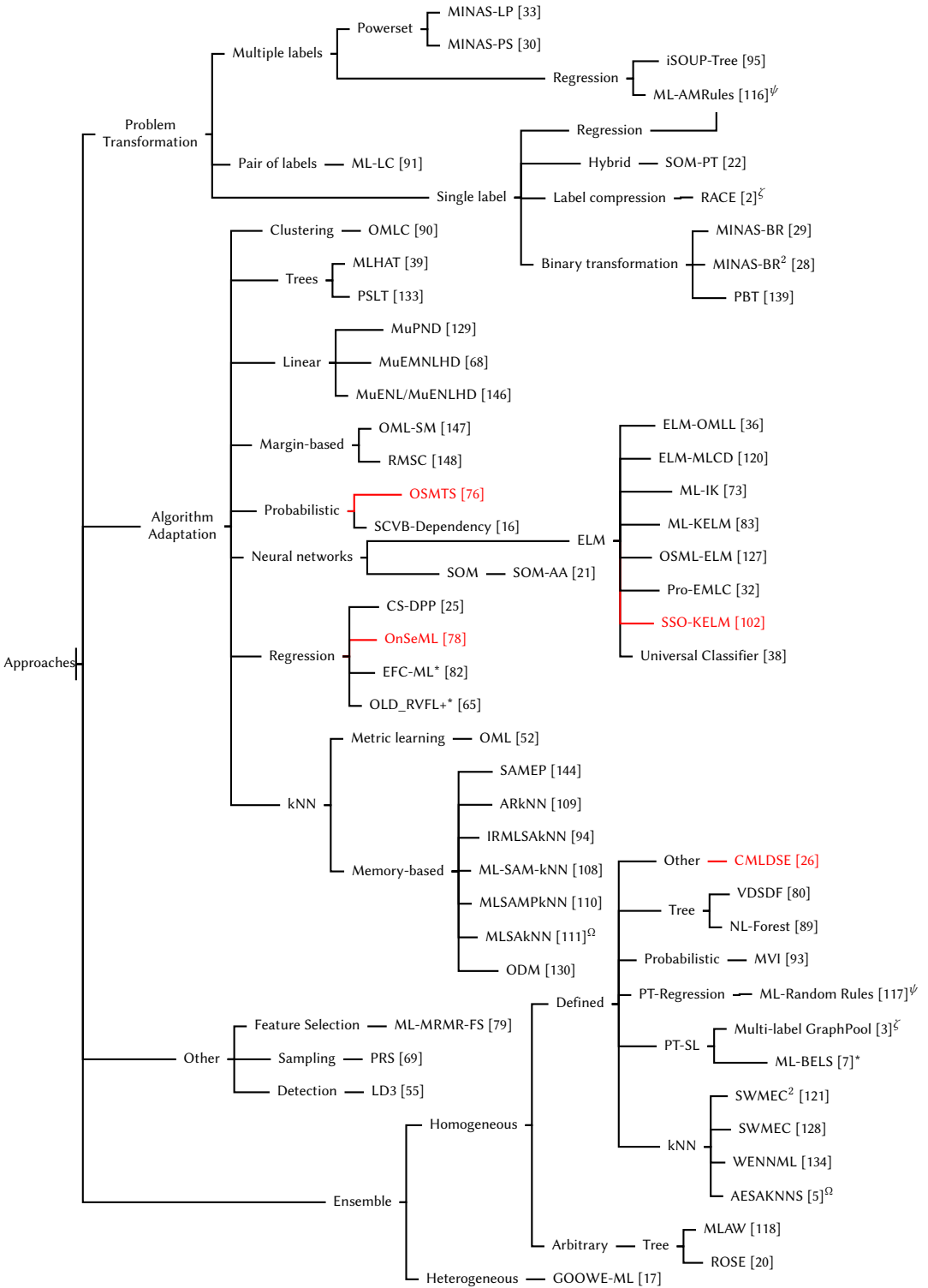

To elicit the structure of a method, a fundamental question is how it approaches the problem, that is, whether a problem transformation (PT) or an algorithm adaptation (AA) approach is employed. Many methods did not explicitly mention the approach. However, many of these indicated the use of, for instance, kNN classifiers. Therefore, for those whose approach is not elicited, we classified them according to the base classifier, given the hierarchy proposed in Figure~\ref{fig:9}.

\subsection{Algorithm adaptation}\label{q1_aa}
ELM, short for Extreme Learning Machine, is a training strategy of feed-forward neural networks that consists of a single hidden layer. This strategy enables a very fast training phase due to the random initialization of the input layer weights and the analytical determination of the output layer weights by means of the Moore-Penrose inverse \cite{Huang2004}. As can be seen in the hierarchy of Figure~\ref{fig:9}, eight methods employ ELMs, which by itself is a strong sign of their power in data streams. SSO-KELM~\cite{Qiu2022}, ML-KELM~\cite{Luo2022}, ML-IK~\cite{Kongsorot2020}, and ELM-MLCD~\cite{Tang2024} make use of a kernel function, which maps input data into a high-dimensional feature space, leading to faster convergence, better generalization, fewer parameters, and avoidance of dimension disaster \cite{Luo2022}. Concerning the matrix inversion, some methods use alternative techniques to compute it, such as Woodbury in ELM-OMLL~\cite{Du2020} and Cholesky decomposition in ML-KELM.

Self-organizing maps (SOM) are another neural network approach employed, both in an AA and a PT fashion. SOM-AA~\cite{Cerri2022} relies on training a single Self-Organizing Map to deal with all the classes simultaneously, which allows it to consider label dependencies, given that the instances assigned to different labels at the same time are mapped to neighboring neurons in the grid. For the PT approach, see the \textit{\hyperref[q1_pt]{Problem transformation}} subsection.

K-nearest-neighbors is another very popular way to deal with data streams, given that eight papers have employed it. The majority of the methods make different use of K-nearest-neighbors by dealing with memory in distinct ways, such as ODM~\cite{Wang2021} and ML-SAM-kNN~\cite{Roseberry2018}, which work relying on a dual memory strategy; MLSAkNN~\cite{Roseberry2021}, MLSAMPkNN~\cite{Roseberry2019}, IRMLSAkNN~\cite{Nicola2024}, and SAMEP~\cite{Zheng2021} work with Self-Adjusting Memory (SAM); and ARkNN~\cite{Roseberry2023} work by weighting instances (see Question \hyperref[q3]{3}). On the other hand, OML \cite{Gong2020}, while still a K-nearest-neighbor-based method, does not use a fixed metric (such as Euclidean distance or cosine) but learns the distance metric within the stream, performing the search for neighbors in a learned embedding space.

The tree-based methods differ essentially in the type of tree used. MLHAT~\cite{Esteban2024} uses an incremental decision tree based on the Hoeffding Adaptive Tree that works by partitioning the feature space to predict the label set. PSLT~\cite{Wei2021}, on the other hand, relies on a tree that organizes the label space according to its correlations, where there is a binary classifier associated with each leaf node, allowing the emergence of new labels (see Question \hyperref[q4]{4}).

OML-SM~\cite{Zou2023} and RMSC~\cite{Zou2024} both rely on a classifier that dynamically adjusts a margin by means of label causality and label prototype. However, the main difference is that OML-SM focuses on missing labels, while RMSC focuses on weak labels and dealing with distribution change in labels (see Question \hyperref[q3]{3}).

As for the probabilistic methods, OSMTS~\cite{Kumar2023} is considered probabilistic because it focuses on the probability of an instance being related to groups of previously seen instances (micro-clusters) and the probability of co-occurrence of labels. On the other hand, SCVB-Dependency~\cite{Burkhardt2018} uses a probabilistic generative model based on Latent Dirichlet Allocation (LDA) that captures the dependencies between labels and estimates the probability of each label being associated with a document.

CS-DPP~\cite{MinChu2019} and OnSeML~\cite{Li2020} depend strongly on regression in order to work. CS-DPP belongs to the family of algorithms called LSDR (Label space dimension reduction), where it learns a predictor that maps the feature vectors to code vectors in a smaller dimension space; linear regression is used to predict the code vector corresponding to a feature vector. Alternatively, OnSeML builds a local regression model for each data instance based on similarity with neighbors and optimization, where the result of the regression in the low-dimensional space is decoded to produce the final prediction. EFC-ML~\cite{Lughofer2022} is a fuzzy classifier based on a multi-output architecture, where for each class, a separate consequent hyper-plane is set, addressing the classification as a binary regression, and the parameters are learned by fuzzily
weighted least squares and Lasso-based regularization. Lastly, OLD\_RVFL+\cite{Huang2023} is a label distribution method where the weights of the model are learned by optimizing an objective function that aims to minimize the difference between the predicted label distribution and the real one, where the distribution of labels is a regression value with membership.

MuENLHD~\cite{Zhu2018}, MUENL~\cite{Zhu2018}, and MuPND~\cite{Wang2019} employ a linear classifier of the form $h_i(\textbf{x}) = \mathrm{sign} (\textbf{w}_i^T \textbf{x} + b_i)$, where $\textbf{w}_i$ is the weight vector and $\textbf{b}_i$ is the bias (scalar), and the learning consist of optimizing these parameters. While MuPND and MuENL do not differ in their classification process (they differ essentially in their concept evolution technique, see Question \hyperref[q4]{4}), MuEMNLHD~\cite{Kanagaraj2023} employs a dimensionality reduction, so the weights are calculated in this new dimension space.

OMLC\footnote{Unofficial name} is a classification method based on weighted clustering that holds mature and immature clusters. The clusters are described by the weighted linear sum of the features, the weighted square sum of the features, and the weight of the clusters. The difference between the mature and the immature is in the weight of the cluster, where the mature has a weight greater than a threshold. Only the mature clusters are used to assign labels to new samples.

\subsection{Problem transformation}\label{q1_pt}

Single-label approaches are mainly devoted to using, in some way, a binary transformation. MINAS-BR~\cite{CostaJunior2019-BR}, MINAS-BR$^{2}$~\cite{Costa2023}, and PBT~\cite{Yildirim2024} are of such kind. While the MINAS methods rely on binary relevance (check section \ref{pt}), PBT orders labels with PCA and transforms the label vectors into integers. RACE~\cite{Ahmadi2018}, on the other hand, is based on label compression and works by compressing the original label space into a smaller space with random projection and decoding the output, aiming to train fewer binary relevance classifiers in these compressed spaces. Finally, SOM-PT~\cite{Cerri2021} consists of training a Self-Organizing Map (SOM) for each label, where the classification is given by the combination of the output of the neurons in the grid.

Multiple-label approaches have the ability to deal with many combinations of labels at once. MINAS-LP~\cite{Costa2017} and MINAS-PS~\cite{CostaJunior2019-PS} rely on the classical Label Power Set strategy, differing in that MINAS-PS actually employs the Pruned Sets variation (check section \ref{pt}). Even though iSOUP-Tree~\cite{Osojnik2016} is based on trees, its fundamental idea is transforming the MLC problem into a multi-target regression (MTR) problem, predicting a numerical value for every label, followed by a binarization according to thresholds. ML-AMRules~\cite{Sousa2016} also transforms the MLC problem into an MTR problem, essentially differing from iSOUP-Tree as the final model is a rule set, where each rule holds a multi-label classifier, which can be a linear or output-mean predictor, chosen adaptively. Also, ML-AMRules actually has three modes, being local (which fits into \textit{single label} approach), global, and subset (fits into \textit{multiple labels} approach), hence being classified in the hierarchy as belonging to both approaches.

Pair of labels explores correlations between pairs of labels. ML-LC~\cite{Nguyen2019_label} is a Bayes-based method that considers pairwise label correlation and relaxes the independence assumption by allowing the dependence of a feature on another feature. The amount of predicted labels is tuned based on label cardinality and Hoeffding inequality.

\subsection{Ensemble}\label{q1_e}

Only one method points to the possibility of mixing different base classifiers (heterogeneous), which is the GOOWE-ML~\cite{Bykakir2018}. It uses spatial modeling of the classifiers' relevance scores to assign them optimal weights. This assignment is achieved by minimizing the Euclidean distance between the combined relevance vector and the ideal vector that represents the ground truth.

The vast majority are homogeneous ensembles, at least on an explicit level. Among those, one can first split between them those that have a defined base classifier and those that employ an arbitrary base classifier. To judge that, we evaluate whether the base classifier used is an essential part of the method (that is, if it is explicit in the name or if it uses an unusual classifier upon which the whole idea of the method is built).

Only MLAW~\cite{Sun2019} and ROSE~\cite{Cano2022} seem to have an architecture independent of the base classifier, which is, therefore, arbitrary. MLAW works by weighting the voting mechanism and tracks label dependency by pruning away rare label sets (similar to PS). ROSE focuses on imbalanced data by means of a sliding window buffer per class (allowing undersampling of the majority classes) and self-adjusting bagging that boosts the exposure to difficult instances from minority classes.

Concerning the defined ones, some are based on AA or PT classifiers that were addressed in this very study. In such cases, one can identify them in the hierarchy of Figure~\ref{fig:9} by checking the Greek symbol super-scripted in the methods (both ensemble and base). Four are based on kNN. SWMEC~\cite{Wang2017} and SWMEC$^{2}$~\cite{Tian2022} differ subtly: SWMEC has a variable size, and SWMEC$^{2}$ has a fixed-size ML-kNN ensemble. WENNML~\cite{Wu2023} is NN-based and dynamically updates the ensemble by means of geometric and diversity weighting methods. AESAKNNS~\cite{Alberghini2022} is an ensemble of MLSAkNN~\cite{Roseberry2021} (check \ref{q1_aa}), which, in turn, relies on Self-Adjusting Memory (SAM) and is therefore fit to concept drift adaptation (see Question \hyperref[q3]{3}).

The BR-based ones, that is, Multi-label GraphPool~\cite{Ahmadi2019} and ML-BELS~\cite{Bakhshi2024}, differ essentially in their base classifiers: the first uses RACE (focused on label space compression, check \ref{q1_pt}), the latter uses BELS (Broad Ensemble Learning System \cite{Bakhshi2023}, a lightweight neural network focused on feature mapping).

ML-Random Rules~\cite{Sousa2018} is an ensemble of ML-AMRules (check \ref{q1_pt} that weights the multiple predictions of its base classifiers, therefore being a regression-based model.

MVI~\cite{Nguyen2019_variational} is an ensemble of VIGO (Online Variational Inference for multivariate Gaussians) classifiers. Being so, VIGOs \cite{Nguyen2018-vigo} are Bayesian classifiers that use variational inference to learn multivariate Gaussian distributions for each class.

VDSDF~\cite{Liang2022} is a deep forest of VFDTs that works in a cascade fashion: first, by building the incremental VFDT and, secondly, by building the hierarchical deep forest structure. NL-Forest~\cite{Mu2018}, on the other hand, uses completely random trees and is structured with two cooperative models: an instance-based forest (built on the whole training dataset) and a label-based forest (multiple sub-forests built considering label information).

CMLDSE~\cite{Chu2019} is an ensemble of classifiers trained with COINS \cite{Zhan2017}, which is an algorithm that works by co-training two classifiers induced on two feature subsets obtained by dichotomization, and then refines the model by communicating the pairwise ranking predictions of unlabeled data.

\subsection{Other approaches}

ML-MRMR-FS~\cite{Li2022} is a multi-label streaming data feature selection method derived from MRMR (Minimal-redundancy and Maximal-relevance \cite{HanchuanPeng2005}), where the goal is to select a small subset of features from the original data that are relevant to the target classes and, at the same time, minimize redundancy among the selected features.

LD3~\cite{Gulcan2023} is not a classification method but a multi-label concept drift detector \textit{per se}. Therefore, in order to check its operation mode, check Question~\hyperref[q3]{3}.

PRS~\cite{Kim2020} is a very particular method as it is designed for continual learning. According to \citet{Gomes2022}, both \textit{data streams machine learning} and \textit{continual learning} learn from incremental data, which likely follow non-stationary distributions and differ essentially by the fact that continual learning usually deals with computer vision problems, mainly employing neural networks and trying to reduce catastrophic forgetting. PRS uses ResNet101 as the NN classifier and uses the replay-based strategy to handle catastrophic forgetting, which stores past samples and "replays" by reusing them, interleaved with novel ones. PRS defines the target proportion of classes in memory by keeping frequency statistics of the running class and then decides if a new sample enters the memory by computing the probability of the minority classes' biased sampling. If the memory is full, the example that most pushes the memory towards the target partition is selected to be removed.

\subsection{Semi-supervised}

The four methods highlighted in red in the hierarchy of Figure~\ref{fig:9}, SSO-KELM, OnSeML, OSMTS, and CMLDSE, have an additional characteristic: they are semi-supervised. Each method employs a technique to deal with the partial absence of labels.

SSO-KELM adapts an offline semi-supervised technique, the SS-ELM \cite{GaoHuang2014}, dealing with the partial absence of labels by making use of a manifold regularization aided by the construction of a similarity graph over both labeled and unlabeled data. The use of a kernel function contributes to a good data representation, taking into account the structure present in the unlabeled instances.

OnSeML, as previously mentioned (check \textit{\hyperref[q1_aa]{Algorithm adaptation}}), relies on the similarity with neighbors to build a local regression model. The method uses both labeled and unlabeled instances in the neighborhood but considers the regression functions learned for the neighbors as pseudo targets, which helps to deal with the unlabeled instances.

OSMTS, as a method that works with micro-clusters, deals with unlabeled data by computing the probability of pertaining to existing micro-clusters, given by the cluster popularity, single-term space similarity, and semantic space by co-occurrence of terms.

CMLDSE is a co-training method, which initially trains two classifiers with the labeled data (with dichotomization in COINS, check \textit{\hyperref[q1_e]{Ensemble}}). Then, these classifiers are used to predict the labels of the unlabeled instances, and the highly trusted labeled instances by one classifier are added to the other classifier's dataset as if it was labeled by an oracle for retraining, followed by several repetitions until it reaches a stop criterion.

\subsection{Baselines}

\begin{figure}[hbtp]
\centering
\includegraphics[scale=0.85]{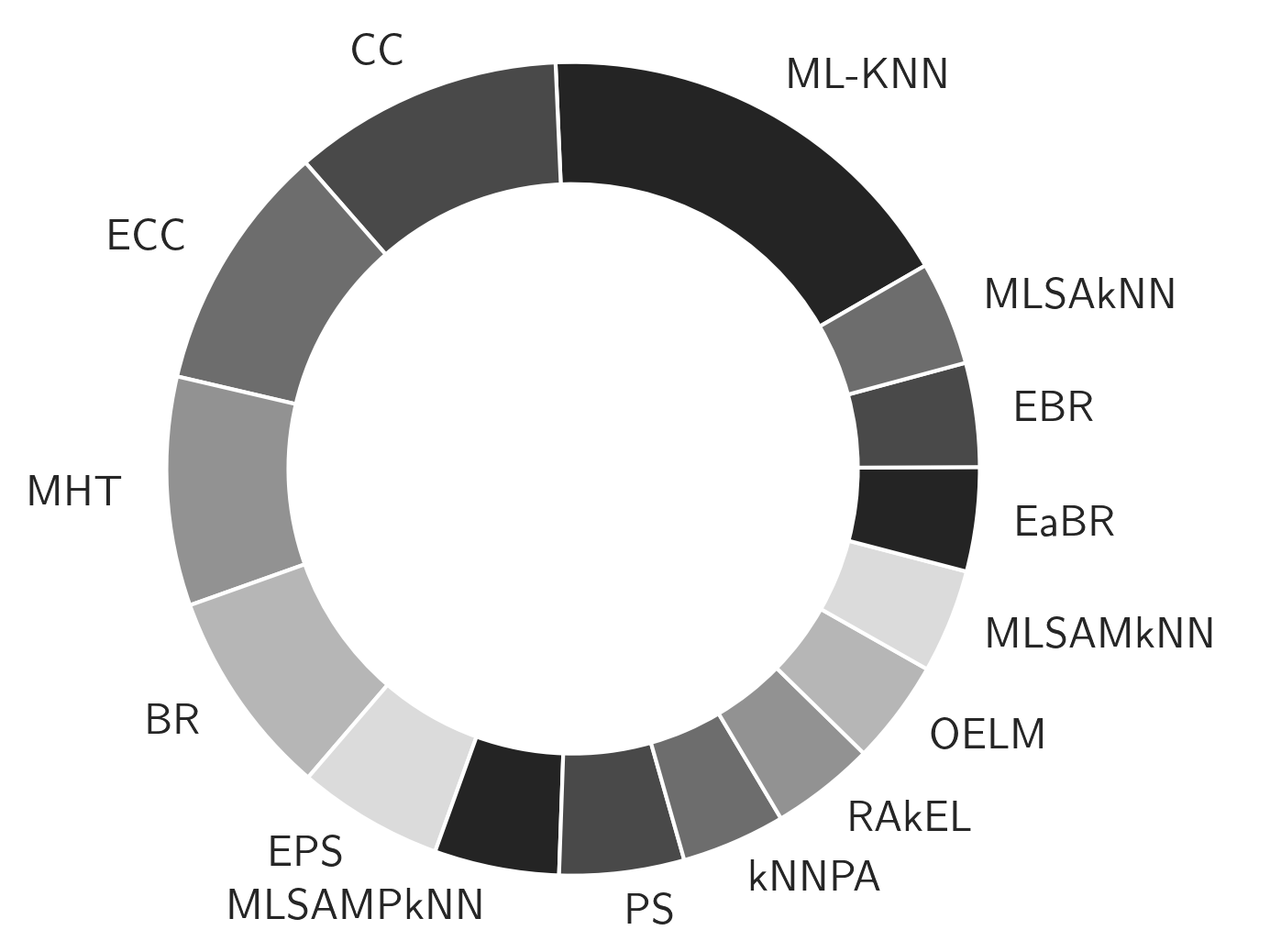}
    \caption{Top 15 most compared baselines.}
\label{fig:10}
\end{figure}

Still, concerning the structure, one can consider which baselines are usually compared to. Figure~\ref{fig:10} depicts the methods most used as comparison baselines. Note that MLSAMPkNN~\cite{Roseberry2019} and MLSAkNN~\cite{Roseberry2021}, studies that are addressed in this review, are among them.

\section{Q2: How does it handle label latency?}\label{q2}

As mentioned in Section \ref{sec1}, ground truth label latency is a poorly addressed theme in data streams. This is an important theme because ground truth is essential in updating a classifier model, and in many real scenarios, access to ground truth is impossible at the same time as the arrival of the data to be classified. Of all the 58 investigated studies, only six considered the possibility of label latency. The six studies considering label latency considered the infinite latency case (as defined in Section \ref{subsec2.3}). No studies concerning delay in feasible time were found. It is worth noting that four of the 58 studies considered a semi-supervised scenario: the total absence of labels in a portion of the data.

It stands out that handling infinite delayed labels in the referred period seems to be a quest of very few researchers, as five of the six papers had the authorship of Cerri, Gama, Faria, and Costa Junior. It comes as shocking news to discover that this problem is, so far, so poorly addressed, given that it reflects a clear real-world problem.

In MINAS-BR \cite{CostaJunior2019-BR}, MINAS-BR$^{2}$~\cite{Costa2023}, and MINAS-PS \cite{CostaJunior2019-PS}, the authors propose the use of k-means to build micro-clusters as decision models for each class so that new data is classified according to the closest micro-clusters. They differ as the first two methods generate the micro-clusters based on subsets for each training class (BR, check section~\ref{pt}), while the latter generates the micro-clusters based on subsets for each frequent label set (PS).

In MuPND \cite{Wang2019}, even though it does not explicitly call itself a method that deals with infinite latency, the test procedure is conducted without access to ground truth labels. Updating the classifier consists of introducing a latent variable that estimates the true label assignment of each instance in the test dataset. Then, the method learns the latent label assignment and the classifier that best fits the data.

In SOM-PT \cite{Cerri2021}, self-organizing maps are built for each class so that the winner neuron (closest to the input) has its weight vector updated according to the information of the previous weights, that is, in an unsupervised fashion. In SOM-AA~\cite{Cerri2022}, the unsupervised weight vector update goes the same way, differing in that there is only one self-organizing map built upon the labeled dataset. The SOM methods represent a potential improvement over the MINAS ones, as there is no need to define \textit{k}: it is replaced by the size of the neuron grid. However, the proposals do not treat concept evolution.

All the described methods share an underlying distance-based nature. This suggests that future work on the infinite label latency problem could explore other methods of the same nature.

\section{Q3: What is the scheme to handle concept drift detection?}\label{q3}

Concept drift detection is the problem that the community has been paying more attention to in the data stream scenario. 32 out of the 59 studies were concerned with dealing with the intrinsic changes of data flowing in time.

\begin{figure}[hbtp]
\centering
\includegraphics[scale=0.6]{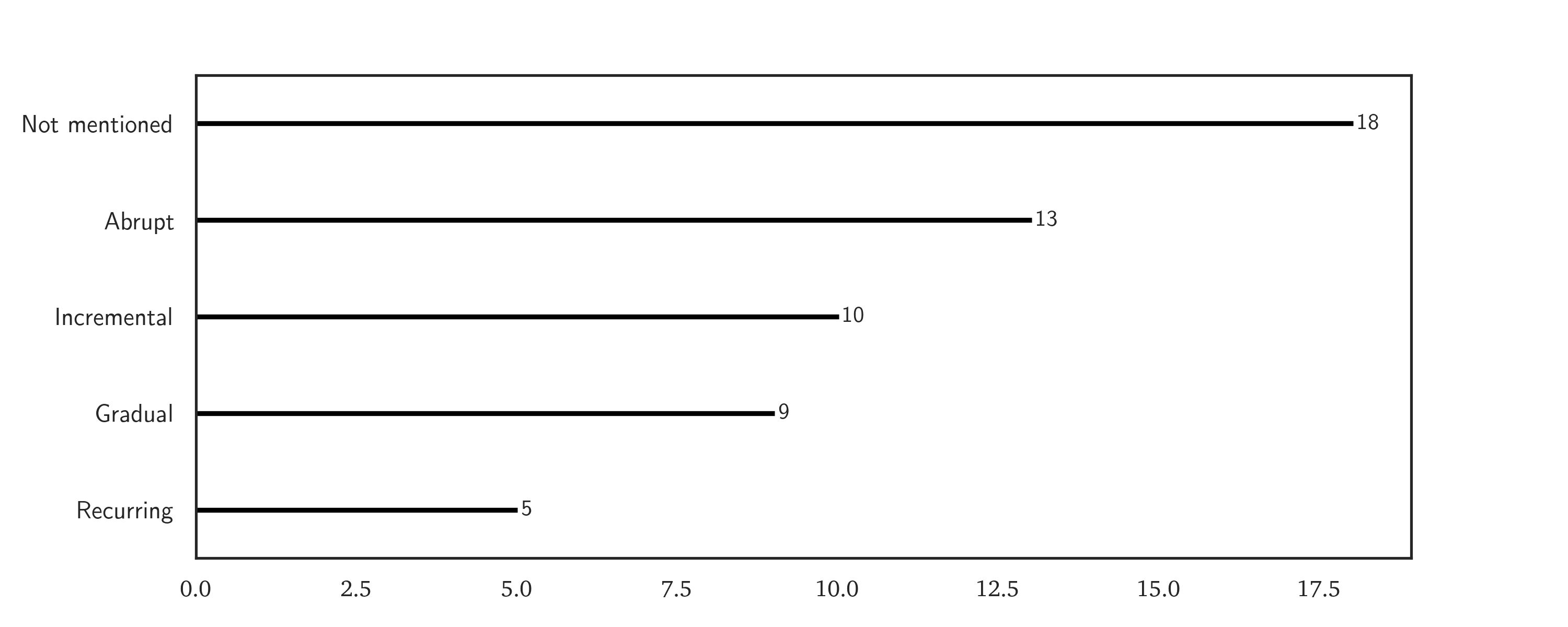}
    \caption{Number of papers that are able to deal with each type of drift.}
\label{fig:11}
\end{figure}

One can understand concept drift according to the patterns of drift. The types of drift were already discussed in Section \ref{drift}. The number of methods dealing with every type of drift is illustrated in Figure~\ref{fig:11}.

The methods to handle concept drift can also be grouped according to the strategy used to adapt to it or detect it. The proposed groups are window-based, ensemble-based, weight-based, decay-based, and hybrid.

\subsection{Window-based}\label{window}

The underlying idea of window-based strategies is to consider snapshots of the streaming data that evolve together with the arrival of instances. The main window-based strategy is to address it with sliding windows, where old instances are discarded, and the most recent are kept and valued equally (not weighted). The window size can be fixed, such as in \cite{Wang2021, Li2022, Liang2022}, or adaptive (dynamically adjustable), as in \cite{Roseberry2018, Roseberry2019, Zheng2021, Roseberry2021}.

Amongst those adaptive size sliding window methods that do not have a detector, the strategy consists of using a short-term memory (STM), which is a dynamic sliding window that adapts itself by constantly evaluating different sub-window sizes and electing the window size with the best score. ML-SAM-kNN \cite{Roseberry2018} uses STM and computes the best score with hamming score (see Section \ref{eval}), while MLSAMPkNN \cite{Roseberry2019}, MLSAkNN \cite{Roseberry2021}, and SAMEP \cite{Zheng2021} also use STM but compute it with subset accuracy (see Section \ref{example-based}).

Amongst those adaptive sliding window methods that have a detector, the strategy is to use ADWIN, short for \textit{ADaptive WINdowing}, which is an algorithm that shrinks the window size when data changes and increases it when there is no change \cite{Bifet2007}. Three methods employ ADWIN as a drift detector: AESAKNNS, ROSE, and WENNML. All these methods rely on ensemble classifiers, which are naturally adaptable to deal with concept drifts, as different classifiers in the ensemble may have learned different aspects of the data.

AESAKNNS \cite{Alberghini2022} uses MLSAkNN \cite{Roseberry2021} as a base classifier, which, as already described, adapts to concept drift by means of STM. AESAKNNS employs a collection of ADWIN detectors to keep track of each classifier. When a drift is detected, additional classifiers are trained to learn the new concepts. ROSE \cite{Cano2022} works pretty much similarly to AESAKNNS, except for the base classifier, which is VFDT. 

WENNML \cite{Wu2023}, a nearest neighbors-based method, also goes in the same way, where ADWIN detects the given warning and leads to the generation of a new data block with the warned instances to train the passive candidate ensemble classifiers (PEC), as opposed to active candidate ensemble classifiers (AEC), which are not trained with warned instances.

Concerning fixed-size sliding windows, VDSDF \cite{Liang2022} uses a simple parametric fixed sliding window. ODM \cite{Wang2021}, on the other hand, also uses a short-term memory, but here, it is a fixed-size window (also parametric) that works in a FIFO fashion coupled with a long-term memory based on reservoir sampling. ELM-MLCD \cite{Tang2024} splits the data stream into equally-sized data blocks, which are partitioned into adjacent windows and then, given a threshold, detects concept drift by evaluating the mean and variance of the instance over the windows.

Two other methods employ a divergence drift detector: ML-MRMR-FS and MLAW. ML-MRMR-FS \cite{Li2022} uses a fixed sliding window strategy and makes use of a divergence drift detector. It computes the divergence between two adjacent data chunks by means of the harmonic mean between the vertical and horizontal distances, and compares it to a threshold to determine whether the label distribution changed or not. MLAW \cite{Sun2019}, on the other hand, uses a double window strategy, where one window is kept for positive examples and another for negative examples in order to deal with unbalanced classes. This method couples it with a concept drift detector, namely Jensen–Shannon Divergence, a symmetric and bounded variant of Kullback–Leibler (see Section \ref{eval} and Table \ref{tab:13}).

\subsection{Weight-based}

This strategy is inherent to neural network methods, as updating the weights of neurons while training is at its very essence. While in SOM-PT~\cite{Cerri2021} and in SOM-AA~\cite{Cerri2022}, the weights of the neurons are constantly updated to absorb the new data distribution, in ML-KELM \cite{Luo2022}, there is the constant update of the weights in its example-incremental learning aspect (E-IL), in contrast with its class-incremental learning (C-IL) aspect (see Question \hyperref[q4]{4}).

An exception is SWMEC \cite{Wang2017}, an ensemble method that relies on the advantage that the different classifiers in the ensemble may have learned different aspects of data, as already explored in Section \ref{window}. In this method, the classifiers are trained with different chunks, and each classifier has a weight, which is updated as new data arrives. As the weight of a classifier gets too low, it becomes obsolete and is replaced by a better-suited one trained in more recent instances.

Other weight-based approaches involve procedures that involve decay functions or fading factors, where the instances are weighted, and their influence on the current concept changes over time. ML-LC and OMLC \cite{Nguyen2019_label, Nguyen2019} both use a decay mechanism where the weight of each sample declines as it gets older. However, ML-LC uses a function that decays data constantly, while OMLC uses a function that decays data exponentially.

\subsection{Hybrid}

The MINAS methods \cite{Costa2017, CostaJunior2019-BR, CostaJunior2019-PS, Costa2023} mix a window-based strategy with a micro-cluster novelty detection mechanism. They consist of using a fixed-size window to evaluate performance, where the instances that do not pertain to any of the available classes are sent to a short-term memory. When this short-term memory is full, new micro-clusters are created and validated. However, the MINAS-BR~\cite{CostaJunior2019-BR} and MINAS-BR$^{2}$~\cite{Costa2023} go a step further, checking whether these new micro-clusters actually represent a concept drift or an entirely novel class (see Question \hyperref[q4]{4}).

ARkNN~\cite{Roseberry2023} uses a fading window where, in spite of a sliding window, there is a fading factor $\delta$ to assign different significance given the age of the information, where newer data have a greater influence on prediction. So, ARkNN deals with concept drift with a hybrid of window-based and weight-based strategies. The method also mixes a pruning strategy to remove instances that are not yet old enough to be removed by their age, but no longer contribute significantly to the prediction.

OSMTS~\cite{Kumar2023} bets on employing different mechanisms to deal with each type of drift. OSMTS is an application method for text that uses evolving micro-clusters to keep the subspace of terms for each label. When dealing with gradual change, it uses a triangular time function to remove outdated terms. When dealing with abrupt change, it uses the Chinese Restaurant Process (CRP) to create new micro-clusters. There can also be the merging of micro-clusters.

ML-BELS~\cite{Bakhshi2024} detects concept drift by mixing ensemble learning (adapts to drift modifying the composition of the ensemble), addition, removal based on accuracy (examples with accuracy lower than a threshold are removed from the active ensemble and are moved to a pool, which can be retrieved to the learning process) and dynamic weighting (incorporates label dependencies which reflect the changes in case of drift).

\subsection{Other}

Multi-label GraphPool~\cite{Ahmadi2019} handles concept drift using a two-stage drift detection method, where the first stage consists of monitoring the performance of the current concept and, given a significant drop, the second stage is activated. This second stage compares the decoding matrix of the new batch of data with the existing concepts in the pool. Based on this comparison and a user-defined threshold, the method can update an existing concept, merge multiple concepts, or add a new concept to the pool.

ELM-OMLL \cite{Du2020} deals with a special type of drift, namely \textit{changes in data distribution with labels} (CDDL), where the changes in data distributions happen dynamically with the labels of incoming examples. The adaptation to CDDL relies on an objective function based on label ranking and a fixed threshold that makes the model robust to changes in data distribution.

ML-AMRules~\cite{Sousa2016} and ML-Random Rules~\cite{Sousa2018} use Page-Hinkley~\cite{Page1954}, which is a detector of significant mean changes in the sequence of data by means of the cumulative sum.

RMSC~\cite{Zou2024} deals with distribution changes in labels (CDL) by means of instance-specific online thresholding, so that this threshold allows it to adjust to label cardinality changes.

MLHAT~\cite{Esteban2024} uses ADWIN, but the concept drift module cannot be considered window-based. An ADWIN detector is employed in each node of the tree and in potential background sub-trees, where there is a replacement of the actual sub-tree by a background sub-tree by means of the Hoeffding bound. Hence, ADWIN is used to compare sub-portions of a variable-size window (given by the expansion of the tree).

LD3~\cite{Gulcan2023}, as said above, is a multi-label concept drift detector \textit{per se}. As an unsupervised approach, the strategy is to monitor dependencies between the predicted labels. This relies on two sliding windows of predicted labels, one for old data and another for new data, followed by the creation of co-occurrence matrices to identify which labels occur together. Local rankings based on each label's co-occurrence frequency are created and then aggregated into two global rankings (old and new data). Finally, the correlation between two global rankings is computed, where a low correlation is signaled as concept drift.

\section{Q4: What is the scheme to handle concept evolution detection?}\label{q4}

Another poorly addressed issue is concept evolution. Of all the 52 investigated studies, only eight considered the possibility of concept evolution. Neither of them considered the possibility of a recurring class.

Concept evolution, as elicited in Section \ref{novelty}, can involve a novelty detection procedure, as well as concept drift. As such, it is pertinent to investigate whether the implementation of a concept evolution detection method is accompanied by a concept drift detection method. To that purpose,  MINAS-BR, MINAS-BR$^{2}$, ML-KELM, and OSMTS \cite{CostaJunior2019-BR, Costa2023, Luo2022, Kumar2023} implemented both concept evolution and concept drift methods, delivering a broader novelty detection routine.

Pro-EMLC~\cite{Dave2016} identifies new labels implicitly when the size of the output tuple increases during sequential learning, indicating the presence of a new class. Then, the method adapts its output weight matrix to accommodate the new label.

MINAS-BR~\cite{CostaJunior2019-BR} evaluates when the short-term memory is full and generates micro-clusters when this happens. It calculates the Euclidean distance between the centroids of the micro-clusters, the cardinality of the data, and a threshold factor to define whether these new micro-clusters represent a concept drift of an already existing class or represent an entirely novel class.

The scheme to handle concept evolution in MINAS-BR$^{2}$~\cite{Costa2023} works pretty much the same way as MINAS-BR~\cite{CostaJunior2019-BR}, with the difference that it takes away the threshold factor and uses the Bayes' rule to calculate the probability of a new micro-cluster belonging to each class, considering the correlations between the classes.

MuENL~\cite{Zhu2018} makes use of MuENLForest as the detector. Each MuENLForest is a forest of MuENLTrees, which are binary decision trees built upon original features of the instances and their predicted labels. In turn, each MuENLTree consists of leaves and internal nodes, where the internal ones divide instances based on their distance from two cluster centers, and the leaf defines a ball (frontier defined by mean and radius) to encompass each instance. The concept evolution is detected when an instance falls outside the ball at a leaf node, indicating the arising of a new label.

MuPND~\cite{Wang2019} also employs MuENLForest to detect instances with potential new labels, applies DBSCAN to group them into clusters representing new labels, and then updates the method by creating new classifiers for each new label and fine-tuning the classifiers of the known labels.

CMLDSE~\cite{Chu2019} is another method that makes use of MuENLForest, being composed of 100 MuENLTrees, where each MuENLTree is built over a random subset of a new data chunk and, if a new label is detected, the classifier is retrained to include the new label.

In~\cite{Kongsorot2020}, the authors propose ML-IK and many variations. Two of these variations are equipped with a novelty detector, which relies on a one-class incremental KELM (D-IK) model. D-IK calculates a novelty score for it using the decision function, such that if the novelty score exceeds the threshold parameter, the instance is considered to have a new label and is added to a buffer. When hitting the maximum size, the kernel output matrix and the weight vector are updated to contemplate the new instances and labels.

In ML-KELM~\cite{Luo2022}, also an incremental KELM model, the class-incremental learning (C-IL) aspect is responsible for adding new groups of nodes in the output layer, allowing the novel instances to be classified as pertaining to this newly arrived label.

PSLT~\cite{Wei2021} is a tree-based method where the subset of labels corresponds to the non-leaf nodes, and a binary classifier is trained in the leaf nodes. The method incorporates new labels into the tree structure based on label proximity and updates node classifiers efficiently by leveraging existing classifiers.

MuEMNLHD~\cite{Kanagaraj2023} deals with new labels rising primarily by means of an outlier detector. Then, the instances likely to have a new label are added to a buffer. When this buffer is full, OPTICS (based on DBSCAN) is used to cluster these instances, where each cluster stands for a new label. Finally, the multi-label classifier is iteratively updated to absorb the new labels.

To handle new labels, OML-SM~\cite{Zou2023} sets a prototype for the new label with the feature vector of the first instance associated with it. For subsequent examples with the new label, its prototype is incrementally updated. The conditional probabilities between the new label and other positive labels are also updated. Through this continuous update of the label set, the prototype matrix, and the causality matrix, OML-SM adapts to the evolving label distribution and incorporates new concepts as they emerge.

NL-Forest~\cite{Mu2018} is an application method explicitly concerned with social streams. It determines the presence of novel labels, comparing the average height (average number of partitions to isolate an instance and detect new labels) of the test instances with thresholds. The method also considers two types of novel labels: Absolutely Novel Labels, observed by isolating an instance-based forest, and Partially Novel Labels (PNL), by exploring a label-based forest.

\section{Q5: What is the evaluation strategy?}\label{eval}

Concerning the evaluation procedure, merely eleven studies are explicit about which they use, being prequential. Even though most papers do not give details about it, it seems that the most common strategy is to run the whole experiment and then evaluate it in the end, as a batch setting.

\begin{figure}[hbtp]
\centering
\includegraphics[scale=0.4]{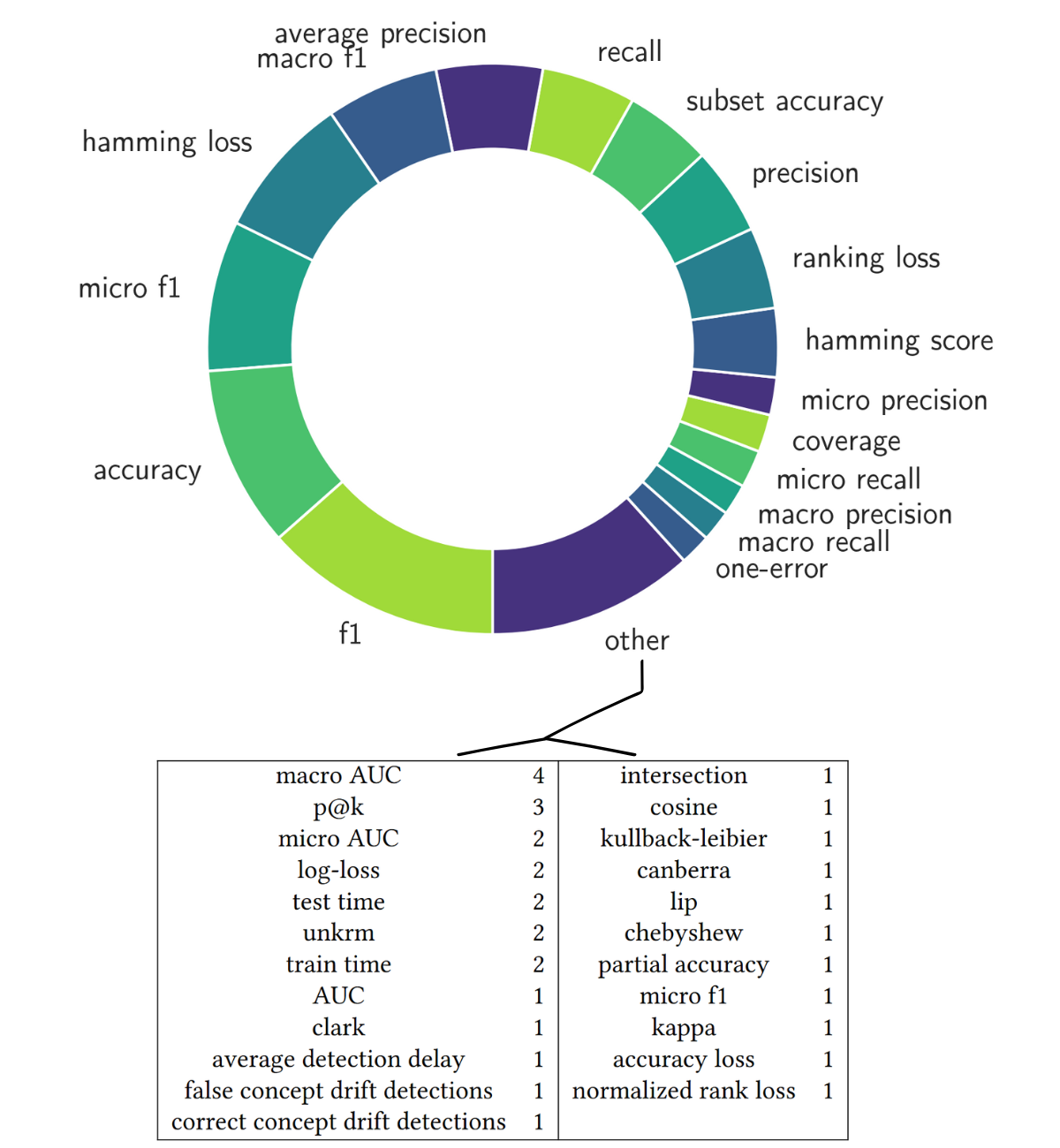}
    \caption{Most used metrics in the analyzed papers.}
\label{fig:12}
\end{figure}

The metrics employed in a multi-label scenario (see Section \ref{ev1}) are far from being consensually established. This is reflected directly in the context of data streams. Figure~\ref{fig:12} depicts the proportionality of the metrics with regard to all metrics used in the papers analyzed.

The most used metric, by far, is the F1-score. 50\% of the papers employed this metric in its example-based form, whereas about 25\% used micro F1 and 20\% used macro F1. This makes F1-score stand out not only as the most example-based metric but as the most label-based metric as well. F1-score is a very convenient metric because it already takes into account two other metrics, Precision and Recall, packing a good amount of information.

All the other example-based metrics are right after, being led by accuracy and hamming loss. Besides hamming loss, a metric appeared that had not yet been explored: hamming score. To understand it, it is enough to look at the equation $HS = \frac{1}{pq}\sum_{i=1}^{p} \sum_{j+1}^{q} \mathcal{I} [\hat{\mathbf{y}}_{j}^{(i)} = \mathbf{y}_{j}^{(i)}]$ and recall that the only difference to HL as exposed in Table \ref{tab:02} is the change of $\mathcal{I}$, as here it is the interpretation of the equality test. That is, in contrast to HL, what is being measured is the number of correct predictions.

The greater use of example-based metrics in the analyzed papers is congruent with the very incremental nature of data streams. The label-based metrics demand averaging for each label and all examples. In a streaming context, where the number of instances can become huge, or even where there might be forgetting mechanisms, obtaining the average can be unfeasible.

If we closely check the chart, it is clear that almost every metric appearing (except coverage) is the regular metric for \textit{batch} multi-label classification task, as listed in Table \ref{tab:02}. Coverage, which was not explained previously, is a ranking-based metric that measures the depth of the array of predicted labels in order to include all relevant labels pertaining to an instance, that is, if the model can cover the whole set of relevant labels for instance~\cite{STidake2018}. According to \cite{Herrera2016-metrics}, it sums up to $\{ \frac{1}{n} \sum_{i=1}^{n} \displaystyle arg~max_{y \in Y_i} \langle rank(x_i,y) \rangle - 1 \}$.

The slice named as ``other'' in the pie chart includes metrics that are not usual or that are specific to the context of the method they were applied. In the next paragraphs, these less-used metrics will be explored, as they were not yet addressed in Section~\ref{ev1}. AUC stands as the most used metric in this slice. To understand the Area Under the ROC Curve, one must first dive into what the Receiver Operating Characteristic (ROC) curve is. In essence, the ROC curve is a curve obtained by plotting many threshold values in a 2D space of false positive rate and true positive rate. That is, when looping from 0 to 1, if considering the threshold 0.4, we get a 0.4 false positive rate and a 0.8 true positive rate, we would have the Cartesian coordinate (0.4,0.8). Bearing it in mind, in multi-label contexts, AUC can be computed in example-based or label-based flavors, as seen in Table \ref{tab:12}. They are all based on some form of averaging: example-based AUC averages over all instances, macro-AUC averages over all labels, and micro-AUC averages over the prediction matrix~\cite{Wu2017}.

\begin{table}[ht]
\caption{Multi-label AUC}
\begin{center}
\begin{tabular}{c}
    \toprule
    \(\displaystyle example-AUC = \frac{1}{m}\sum_{i=1}^{m} \frac{\{ (u,v) \in Y_{i\cdot}^{+} \times Y_{i\cdot}^{-} | f_u (x_i) \geq f_v (x_i) \}}{|Y_{i\cdot}^{+}| |Y_{i\cdot}^{-}|} \)\\[1.3em]
    \(\displaystyle macro-AUC = \frac{1}{l}\sum_{j=1}^{l} \frac{\{ (a,b) \in Y_{\cdot j}^{+} \times Y_{\cdot j}^{-} | f_j (x_a) \geq f_j (x_b) \}}{|Y_{\cdot j}^{+}| |Y_{\cdot j}^{-}|} \)\\[1.3em]
    \(\displaystyle micro-AUC = \frac{\{ (a,b,i,j) | (a,b) \in Y_{\cdot i}^{+} \times Y_{\cdot i}^{-},  f_i (x_a) \geq f_j (x_b) \}}{(\sum_{i=1}^{m}|Y_{i\cdot }^{+}|) \cdot (\sum_{i=1}^{m} |Y_{i\cdot}^{-}|)} \)\\
    \\[-0.5em]
    \hline
    \multicolumn{1}{p{10cm}}{Notes. $Y_{i \cdot}$ is the \textit{i}th row vector and $Y_{\cdot j}$ is the \textit{j}th column vector of the label matrix. $Y^{+}_{i\cdot} = \{j | \mathbf{y}_{ij} = 1 \}$; $Y^{-}_{i\cdot} = \{j | \mathbf{y}_{ij} = 0 \}$. $Y^{+}_{\cdot j}$ (or $Y^{-}_{\cdot j}$) stands for the index set of positive (or negative) instances of the \textit{j}th label. $|\cdot|$ is the cardinality of a set. $f_j (x_i)$ is the predicted value of $\mathbf{y}_{ij}$ \cite{Wu2017}.}\\
    \hline
\end{tabular}
\end{center}
\label{tab:12}
\end{table}


As AUC implies averaging, it is a troublesome metric to be used in streaming contexts. As a result, only ROSE \cite{Cano2022} is explicit in using it in a prequential fashion. Prequential AUC \cite{Brzezinski2017} is an example-based manner to compute AUC, but without the use of the average as in Table \ref{tab:12}. It uses a sliding window to focus only on the most recent data and a red-black tree to keep the scores ordered. Given a new instance, the classifier score is inserted into the window and into the tree, removing the oldest score if the window is full. The summation of the positive examples that precede the negative examples is the AUC, which is normalized by the total number of possible pairs.

The authors of ML-KELM \cite{Luo2022}, despite not being explicit in defining what type of AUC they are using, make clear that they are averaging over all class labels, which makes it the macro-AUC. VDSDF \cite{Liang2022}, PSLT \cite{Wei2021} and SCVB-Dependency \cite{Burkhardt2018} also make resort to macro-AUC. PSLT \cite{Wei2021} and SCVB-Dependency \cite{Burkhardt2018}, besides employing macro-AUC, also employ micro-AUC. None of these methods specifies which evaluation procedure they employ.

It is important to notice that, in spite of being a batch evaluation, it is not a holdout evaluation. As a matter of fact, no methods explained the use of holdout, nor was it detected. What happens in those cases (which include the methods elicited in the previous paragraph) is that a sort of ``round'' of execution is performed. Then, in the end, the metric is computed for the whole stream.

Precision at top-k (p@k) is a ranking-based metric that computes the precision for the top-K predicted labels. OnSeML \cite{Li2020}, PSLT \cite{Wei2021}, and ML-KELM \cite{Luo2022} employ this metric, such that OnSeML makes use of p@1, p@3, and p@5, PSLT of p@1, and ML-KELM of p@1 and p@3. The authors of PSLT \cite{Wei2021} formally define p@k as:  $\{ p@k = \frac{1}{n} \sum_{i=1}^{n} \frac{|Y^{+}_{i\cdot} \cap \top_{k} (f(x_i))|}{k} \top_{k} (f(x_i)) = \{ j|f^{j} (x_i) \in \top_{k} (f^{1} (x_i), \dots, f^{m} (x_i)) \} \}$.

iSOUP-Tree~\cite{Osojnik2016} and MLAW~\cite{Sun2019} employ the logarithmic loss, which they define as a ranking-based metric, with outcome ranging from 0 to 1 and holding the characteristic of being a ``harder'' error punisher. \citet{Sun2019} elicit the log-loss as: $\{ log-loss(\hat{y},y) = -(ln(\hat{y})y + ln(1-\hat{y})(1-y)) \}$.

Pro-EMLC~\cite{Dave2016} computes the Label Introduction Pattern (LIP), which is a modest metric that simply states how many new labels were introduced in the sequential phase of the experiment. As one might recall from Question \hyperref[q4]{4}, Pro-EMLC is a method that deals with concept evolution, hence the different number of labels in the initial phase and in the streaming phase.

MINAS-BR \cite{CostaJunior2019-BR} and MINAS-BR$^2$ \cite{Costa2023} employ the Macro Unknown Rate. This label-based metric measures the evolution in time of the number of unknown examples. In the context of these methods, examples can be classified in an existing class or can be used to update the model with the emergence of a novel class (check Question \hyperref[q4]{4}). In case an example cannot be used in either of these situations, it is considered an \textit{unknown} example. The metric is defined as $\{ UnkRM = \frac{1}{q}\sum_{i=1}^{q}\frac{UNK_i}{N_i} \}$, where $UNK_i$ is the counting of unknown examples from the label $y_i$, $N_i$ is the counting of examples from the same label and $q$ is the amount of labels in the evaluated time window (see \textit{Label-based} at Section \ref{label-based}).

ROSE~\cite{Cano2022} makes use of the $\kappa$ (kappa) metric, which assesses the inter-rater agreement between the distribution of classes and the successful predictions, ranging from -100 (total disagreement) to 100 (total agreement), where 0 is the default probabilistic classification. According to \citet{Cano2019}, $\kappa$ can be defined as: $\{ \kappa = \frac{n \sum_{i=1}^{c} x_{ii} - \sum_{i=1}^{c}x_i.x._i}{n^2 - \sum_{i=1}^{c}x_i . x._i} \cdot 100 \}$, where $n$ is the count of examples, $c$ the count of classes, $x_{ii}$ the count of successful predictions, and $x_i$,$x._i$ are the total counts of rows and columns.

OLD\_RVFL+~\cite{Huang2023} is a method that employs many odd metrics, as it is a Label Distribution Learning method. Namely, there are four distance metrics: Chebyshev distance, Clark distance, Canberra metric, and Kullback-Leibler divergence, and two similarity metrics: Cosine and Intersection. Table~\ref{tab:13} sums them all up~\cite{Cha2007}.

\begin{table}[ht!]
\caption{Label Distribution Learning metrics}
\begin{center}
\begin{tabular}{c|c}
    \toprule
    Chebyshev & \(\displaystyle d_{Cheb} = \max_i |P_i - Q_i| \)\\
    \\[-1em]
    \hline\\[-1em]
    Clark & \(\displaystyle d_{Clk} = \sqrt{\sum_{i=1}^{d} \left( \frac{|P_i - Q_i|}{P_i + Q_i} \right)^{2}} \)\\
    \\[-1em]
    \hline\\[-1em]
    Canberra & \(\displaystyle d_{Can} = \sum_{i=1}^{d} \frac{|P_i - Q_i|}{P_i + Q_i} \)\\
    \\[-1em]
    \hline\\[-1em]
    Intersection & \(\displaystyle S_{IS} = \sum_{i=1}^{d} \min(P_i,Q_i) \)\\
    \\[-1em]
    \hline\\[-1em]
    Cosine & \(\displaystyle S_{Cos} = \frac{\sum_{i=1}^{d}P_i Q_i}{\sqrt{\sum_{i=1}^{d} P_{i}^{2}} \sqrt{\sum_{i=1}^{d} Q_{i}^{2}}} \)\\
    \\[-1em]
    \hline\\[-1em]
    Kullback-Leibler & \(\displaystyle d_{kl} = \sum_{i=1}^{d} P_i \ln \frac{P_i}{Q_i} \)\\
    \\[-1em]
    \hline
\end{tabular}
\end{center}
\label{tab:13}
\end{table}

Interpreting the metrics according to \cite{Cha2007}, one can state:

\begin{itemize}
    \item Chebyshev is a metric that takes into consideration the worst match over the whole label distribution;
    \item Clark is a dissimilarity metric between two vectors in such a fashion that it amplifies the differences as the values are small;
    \item Canberra is clearly very similar to Clark in such a way that it is quite sensitive to small changes near 0;
    \item Intersection quantifies the overlap between two vectors by summing the smallest of the corresponding elements in each vector;
    \item Cosine is a similarity metric that computes the angle between two vectors;
    \item Kullback-Leibler is an entropic metric, as it measures the information loss in approximating two distributions.
\end{itemize}


CS-DPP~\cite{MinChu2019} makes use of two losses not used by any other paper, namely, normalized rank loss and accuracy loss. The normalized rank loss is merely the normalization (that is, fitting into a range of 0 and 1) of ranking loss (see Section \ref{ev1}), as the authors define: $\{ C_{NR}(\mathbf{y}, \hat{\mathbf{y}}) = average_{\mathbf{y}[i] > \mathbf{y}[j]}(\llbracket \hat{\mathbf{y}}[i] < \hat{\mathbf{y}}[j] \rrbracket + \frac{1}{2} \llbracket \hat{\mathbf{y}}[i] = \hat{\mathbf{y}}[j] \rrbracket) \}$. Notice that the normalization is based on the number of possible pairs, making the metric independent of the number of items. On the other hand, the accuracy loss calculates the complement to the fraction of correct positive predictions relative to the total number of cases where the prediction is positive. The authors define it as: $\{ C_{ACC}(\mathbf{y},\hat{\mathbf{y}}) = 1 - \left( \sum_{k=1}^{K} \llbracket \mathbf{y}[k] = +1~\text{and}~\hat{\mathbf{y}}[k] = +1 \rrbracket \right) / \left( \sum_{k=1}^{K} \llbracket \mathbf{y}[k] = +1~\text{or}~ \hat{\mathbf{y}}[k] = +1 \rrbracket) \right) \}$.

In order to evaluate the quality not of the classifier but of the concept drift method itself, the authors of LD3 \cite{Gulcan2022} compute the \textit{average detection delay} (given in terms of samples) and the number of both \textit{false concept drift detections} and \textit{correct drift detections}, which, as they also state, generally overlooked given that there are not enough test data in which they can be assessed.

In the context of a fuzzy multi-label classifier, EFC-ML \cite{Lughofer2022} employs yet another unusual metric: partial accuracy. It is given by summing up, for all instances, the proportion of correctly predicted classes for each instance in relation to the total number of classes, as expressed in: $\{ PA=\frac{1}{KN} \sum_{k=1}^{N} \left| 
\left\{ \mathbb{I}_{\hat{\mathbf{y}}_{c}(k) == \mathbf{y}(k)} = 1 \right| \right\}\}$.


\section{Q6: Which are the limitations and proposals for future work?}\label{sec:q6}


As exposed specifically in each of our research questions, there are some limitations in the selected studies. To sum up, in general, they are: i) lack of proper multi-label streaming datasets, ii) lack of strategies to deal with different ground truth label latencies (both delayed and infinite), iii) lack of strategies to \textit{explicitly} (most studies do not mention) deal with multiple degrees of concept drift (especially recurring concepts), iv) lack of strategies to handle concept evolution (especially recurring labels, which none of the papers did), v) lack of unity and consensus on the best metrics to properly evaluate a multi-label data stream classification task.

The proposals for future works generally follow well-defined lines. Some works (e.g., \cite{Bykakir2018, Ahmadi2018, Costa2023, Venkatesan2017}) propose addressing label dynamics, whether by including mechanisms to deal with changes or experimenting with it in contexts with changes (concept evolution, concept drift, recurring concepts), or by considering label correlations. Other works, such as \cite{Roseberry2018, Nguyen2019_label, Kongsorot2020, Wang2021, Tian2022, Alberghini2022, Li2022} propose improvements in efficiency and scalability. Some works propose changes in evaluation and methodology (use of different data, other metrics), such as \cite{Costa2017, Nicola2024, Wang2019}. There are also proposals concerning exploring new learning scenarios (incomplete data, lack of supervision) (e.g., \cite{Gulcan2023, Zou2024}) and exploring using ensemble techniques \cite{Ahmadi2018, Kanagaraj2023}. Improving how features are represented \cite{Wu2023, Roseberry2023} and the architecture of the model (e.g., \cite{Cerri2021, Lughofer2022}) are constant proposals. Lastly, \citet{Mu2018} propose deeper investigations into mathematical and theoretical foundations.

Some of the previous proposals for future work are already being implemented. Examples are the proposals of \citet{CostaJunior2019-BR} (MINAS-BR) and \citet{CostaJunior2019-PS} (MINAS-PS), which extended the method proposed by \citet{Costa2017}. Also, \citet{Costa2023} (MINAS-BR$^2$) implemented some future directions pointed by \citet{CostaJunior2019-BR}. We can also find some investigations by \citet{Roseberry2019} (MLSAMPkNN) on the directions pointed by \citet{Roseberry2018}, and by \citet{Roseberry2021} (MLSAkNN) extending the proposal of \citet{Roseberry2019}.

\section{Methods proposed for specific domains}\label{sec:appl}

While most of the literature proposes general methods to deal with multi-label data streams from any application, some proposals were originally designed to deal with specific kinds of problems.

NL-Forest~\cite{Mu2018}, for example, is a method that aims to address Social Stream Classification with emerging New Labels (SSC-NL). It deals with a continuous stream of text data, such as tweets and Weibo micro-blogs, and represents text utilizing Word2Vec~\cite{Mikolov2013}.

OSMTS~\cite{Kumar2023} is also designed for texts and represents them with an evolving micro-cluster that consists of many characteristics, such as the number of documents, term frequency of a word, the sum of frequency count of all words in the cluster, word-to-word co-occurrence score
matrix, assigned label of micro-cluster, decay weight, last updated timestamp, and arriving time stamps of the words.

SCVB-Dependency~\cite{Burkhardt2018} deals with texts in a different fashion by topic modeling. The model associates labels and topics with distributions over words, therefore capturing the dependency between labels.

PRS~\cite{Kim2020}, as a continual learning method, focuses on image classification datasets. It performs experiments using ResNet101, RNN-Attention, and Attention Consistency, all well-established image classification architectures, but sticking to ResNet101 as the main one.

\section{Conclusions}\label{sec:conclusion}

As the nature of data has evolved into a streaming context, characterized by the continuous arrival of thousands or millions of records daily, the need for efficient incremental learning algorithms has emerged. This systematic review sought to provide an in-depth analysis of multi-label data stream classification methods, where instances are simultaneously classified into many classes. The contributions include a complete hierarchy of methods, an exhaustive list of their asymptotic complexities, and an exploration of the most recent strategies for dealing with label latency, concept drift detection, and concept evolution detection. Additionally, a detailed discussion of the evaluation strategies adopted and an identification of the main gaps and future directions for research were carried out.

Our findings indicate that the field of multi-label data stream classification is exceptionally dynamic, with methods being developed with the most diverse strategies for dealing with different problems inherent to streaming data. Also, the limitations of our work that may point to future research directions include: i) lack of an appropriate and detailed investigation on methods proposed for specific domains, exploring both the technique and the scenario itself; ii) lack of an appropriate and detailed investigation on the semi-supervised context; iii) lack of a rigorous mathematical evaluation of the asymptotic complexities in Table~\ref{tab:10}; and iv) lack of benchmarking the methods so a real runtime comparison could be obtained (as opposed to Table~\ref{tab:11}).

\bibliographystyle{ACM-Reference-Format}
\bibliography{sample-base}

\end{document}